\documentclass[iicol]{sn-jnl}

\usepackage{graphicx}%
\usepackage{multirow}%
\usepackage{amsmath,amssymb,amsfonts}%
\usepackage{amsthm}%
\usepackage{mathrsfs}%
\usepackage[title]{appendix}%
\usepackage{xcolor}%
\usepackage{textcomp}%
\usepackage{manyfoot}%
\usepackage{booktabs}%
\usepackage{algorithm}%
\usepackage{algorithmicx}%
\usepackage{algpseudocode}%
\usepackage{listings}%
\usepackage{pifont}
\usepackage{amsmath}   %
\usepackage{bbm}       
\usepackage{graphicx}  
\usepackage{makecell}
\usepackage{bm}
\usepackage{afterpage}
\usepackage{multirow}
\usepackage{makecell}
\usepackage{xurl}
\usepackage{hyperref}

\DeclareMathOperator{\MinMaxNorm}{MinMaxNorm}

\newcommand{\red}[1]{\textcolor{black}{#1}}

\theoremstyle{thmstyleone}%
\theoremstyle{thmstyletwo}%

\theoremstyle{thmstylethree}%

\begin{document}

\title[Multi-view Optimal Transport for Multi-view Crowd Localization and Tracking]{Multi-view Optimal Transport for Multi-view Crowd Localization and Tracking}


\author[1]{\fnm{Kaiyi} \sur{Zhang}}\email{zhangky1999@gmail.com}

\author*[1]{\fnm{Qi} \sur{Zhang}}\email{qi.zhang.opt@gmail.com}

\author[1]{\fnm{Jixuan} \sur{Chen}}\email{chen.jixuanstu@gmail.com}

\author[2]{\fnm{Antoni B.} \sur{Chan}}\email{abchan@cityu.edu.hk}

\author[1]{\fnm{Hui} \sur{Huang}}\email{hhzhiyan@gmail.com}

\affil[1]{\orgdiv{Guangdong Provincial Key Laboratory of Visual Media and Multidimensional Intelligence, CSSE}, \orgname{Shenzhen University}, \orgaddress{ \city{Shenzhen}, \country{China}}}

\affil[2]{\orgname{Department of Computer Science, City University of Hong Kong}, \orgaddress{\city{Hong Kong}, \country{China}}}

\abstract{
Multi-view crowd localization and tracking aim to predict the ground-plane positions of all individuals in the scene and estimate their trajectories over time. 
Existing multi-view localization methods typically rely on Gaussian-smoothed density map supervision, whose inherent ambiguity in crowded regions may blur local peaks and degrade localization accuracy. In this work, we introduce a multi-view optimal transport framework that directly supervises point-level predictions on the ground plane, thereby avoiding the limitations of density map regression. 
Specifically, we propose a Mahalanobis distance-based multi-view optimal transport loss (M-MVOT), whose transport cost incorporates view ray direction and object-to-camera distance within a Mahalanobis formulation, with the direction guiding the ellipse orientation and the distance modulating the cost magnitude.
For multi-view tracking, we first extend the proposed localization framework in a tracking-by-detection manner (MVOT-TBD). To further improve performance, we introduce a \red{tracking-specific} optimal transport loss (MVOT-Track) that explicitly supervises inter-frame association. Extensive experiments demonstrate that the proposed optimal transport-based framework consistently outperforms density map-based localization and tracking on multiple multi-view crowd datasets. Code is available at \url{https://github.com/zqyq/MVOT-Track}.}

\keywords{Multi-view, Optimal transport, Crowd localization, Crowd tracking}



\maketitle

\section{Introduction}\label{sec1}

\label{sec:intro}

Multi-view crowd localization and tracking \cite{hou2020multiview, hou2021multiview, song2021stacked, qiu20223d, Teepe_2024_CVPR, ZHANG2026113396} aims to estimate the positions of people on the ground plane in each frame and associate them across time to form trajectories, which can be applied to various scenarios such as crowd analysis, autonomous driving, and public transportation management \cite{ji2026view, anderson2026model}. Multi-view crowd localization fuses multi-camera information via feature extraction and projection of each camera view into a common ground plane, followed by multi-view fusion and decoding. 
Current multi-view crowd localization methods mainly rely on fixed-size Gaussian-kernel density maps as supervision to train the multi-view crowd localization models \cite{hou2021multiview}.
The people locations are further extracted after a local non-maximum suppression (NMS) as a post-processing step on the estimated crowd density maps, where the pixel-wise mean square error (MSE) loss between the predicted and ground-truth density maps is adopted to train the whole model. However, these methods' localization performance is limited by the ambiguity of the density maps in crowded areas, where there may not be clear density peaks over each person for localization. Furthermore, 
multi-view crowd tracking also requires accurate location predictions, as reliable trajectory estimation for all individuals in the scene fundamentally depends on precise localization.


\begin{figure}[t]
\centering

\includegraphics[width=\linewidth]{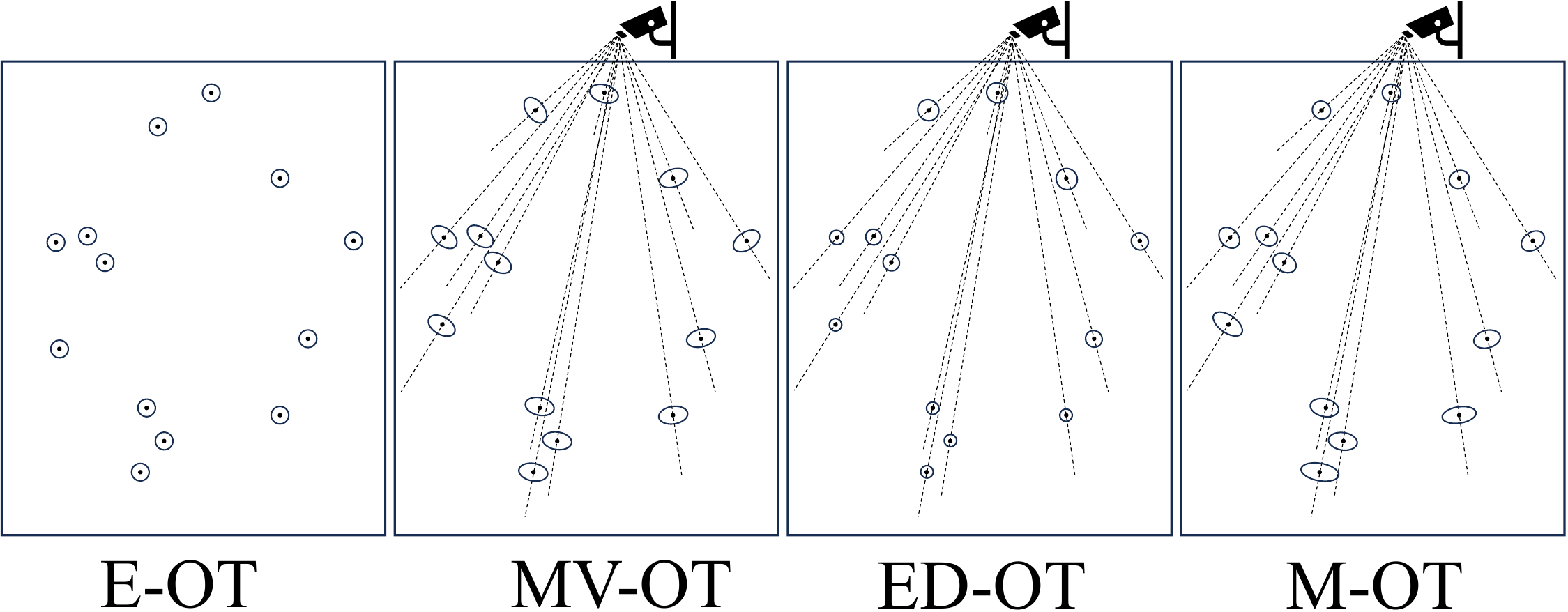}

\vspace{0.5cm}

\setlength{\tabcolsep}{6pt}

\begin{tabular}{lcc}
\hline
Loss  & w/ View Ray & w/ Camera Dist. \\
\hline
E-OT         & \ding{55} & \ding{55} \\
MV-OT       & \ding{51} & \ding{55} \\
ED-OT        & \ding{55} & \ding{51} \\
M-OT & \ding{51} & \ding{51} \\
\hline
\end{tabular}

\vspace{0.3cm}
\caption{The comparison of different optimal transport (OT) losses on the ground plane. The original OT uses Euclidean distance cost (E-OT) and treats all deviations from the ground-truth equally. MV-OT uses the view ray direction to the camera to change the cost function, ED-OT considers the camera distance in E-OT, while M-OT considers both the influence of the view ray direction and the distance to the camera.}
\label{fig:main_idea}

\end{figure}

  To mitigate the problems of Gaussian density map supervision in the single-image crowd localization task, point-supervision methods \cite{ma2021learning,wan2021generalized} based on optimal transport (OT)  
  have been proposed and achieved significant gains in localization performance compared to Gaussian density map methods trained with MSE loss. 
  OT loss directly uses point annotations as supervision and generates more compact density maps \cite{wan2021generalized}.
  However, point supervision for multi-view crowd localization methods has not been explored yet.

  In this paper, we explore using point supervision for multi-view crowd localization and propose a novel Mahalanobis distance-based multi-view optimal transport (M-MVOT) loss. In the M-MVOT loss, the transport cost is defined using the Mahalanobis distance, which adjusts the cost based on the view ray direction between the ground-truth point and the camera, and the distance to the camera. Specifically, the Mahalanobis distance defines elliptical iso-contours of the cost function for a ground-truth annotation, where the ellipse's long- and short-axes are guided by the view ray direction of the camera, and the axes extents are guided by the distance of the point to the camera. 
  
 We first define optimal transport loss for each single camera view.
 In a multi-view crowd localization framework, the projection step of each view causes streaking artifacts of the features along the view ray on the ground plane, which could smear the predicted density map, making localization less accurate. In order to counteract this effect, we penalize predictions that deviate from the ground-truth position more along the view ray. This can be implemented by assigning larger weights in the Mahalanobis distance along the view ray direction. In addition, the object-to-camera distance in each view is used to further adjust the cost, where the locations far away from the camera are more heavily penalized in order to overcome the more severe projection errors there, compared to locations near the camera. We denote this design as M-OT. As illustrated in Fig. \ref{fig:main_idea}, we compare different designs of optimal transport losses, including E-OT, MV-OT, ED-OT, and M-OT. E-OT uses Euclidean distance and considers each point equally, regardless of camera locations. MV-OT uses the Mahalanobis distance based on the view ray direction. ED-OT considers the camera distance in E-OT, and M-OT represents our final design that accounts for both the view ray direction and the camera distance.

  \begin{figure*}[t]
    \begin{center}
       \includegraphics[width=0.98\linewidth]{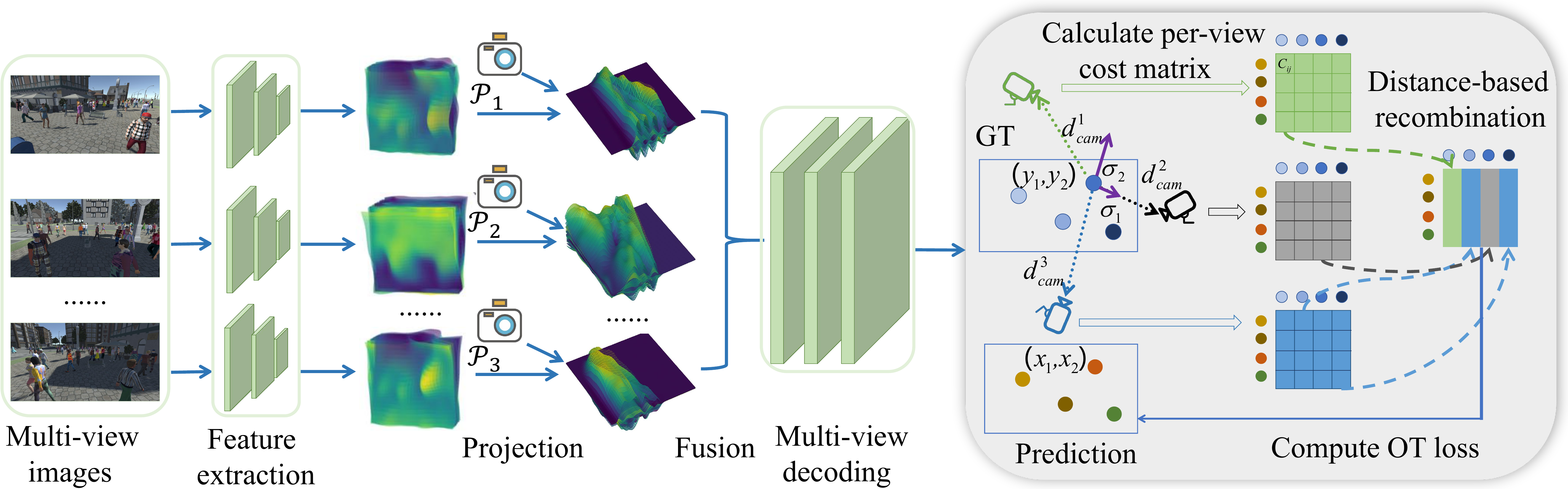}
    \end{center}
    \caption{The model architecture and the proposed M-MVOT loss for multi-view crowd localization. The model consists of feature extraction, projection, and multi-view fusion and decoding. In the proposed M-MVOT, each point's transport cost $\textbf{C}$ is calculated via the Mahalanobis distance instead of the common Euclidean distance \textbf{under the closest camera}, which is \textbf{directed by the view ray and adjusted by the object-to-camera distance}.}

\label{fig:pipeline}
\end{figure*}

  Finally, to consider all the input camera views in the model loss (M-MVOT), we integrate the transport costs (M-OT) over multiple views, by setting each point's optimal transport cost to be calculated with the closest camera. 
  Fig. \ref{fig:pipeline} shows the overall architecture of the multi-view crowd localization model and the proposed M-MVOT loss. The model consists of single-view feature extraction, ground-plane projection, and multi-view fusion and decoding steps. In the proposed M-MVOT loss, the cost for transporting pixels from the predicted density map to the ground-truth points is measured with the corresponding Mahalanobis distance to obtain the transport cost $\textbf{C}$ for the optimal transport problem.

\red{\textbf{Furthermore, we extend the M-MVOT-based localization model to multi-view crowd tracking through two methods, MVOT-TBD and MVOT-Track.}}

\red{First, we develop MVOT-TBD, a tracking-by-detection extension of M-MVOT. Following EarlyBird~\cite{Teepe_2024_WACV}, MVOT-TBD combines a ReID module with a two-stage association algorithm. The ReID module predicts appearance embeddings from the decoded ground-plane features, while the two-stage association algorithm links individuals across frames using their spatial positions and appearance embeddings.}

\red{Although MVOT-TBD is effective, it processes frames independently and does not explicitly optimize temporal association. We therefore introduce MVOT-Track, a tracking-specific OT loss that directly supervises inter-frame associations. The network takes two consecutive multi-view frames as input, adopts the late-to-early temporal fusion strategy~\cite{he2023lef}, and predicts the current-frame localization together with backward motion offsets. These offsets are used to warp the current-frame predictions to the previous frame. MVOT-Track then minimizes the OT loss between the warped predictions and the ground-truth positions at the previous frame, while directly supervising both the transport plan $\textbf{P}^{tra}$ and its marginals. Unlike M-MVOT, whose transport plan matches predicted density with ground-truth points within a single frame, the transport plan in MVOT-Track represents temporal person correspondences across adjacent frames. Its marginal constraints further distinguish individuals who remain in, exit, or newly enter the scene.}

The main contributions of this work are listed as follows.
\begin{itemize}

\item To the best of our knowledge, this is the first exploratory study on using point supervision optimal transport (OT) for multi-view crowd localization and tracking.

\item We propose a novel optimal transport loss M-MVOT for multi-view crowd localization, which adapts the Mahalanobis cost with view ray directions and object-to-camera distances and extends it to multi-view via the closest camera. \red{We further develop MVOT-TBD by integrating a ReID module and a two-stage association strategy, thereby extending the per-frame localization model into a complete tracking-by-detection framework.}

\item To further improve multi-view tracking, we introduce MVOT-Track, \red{a tracking-specific optimal transport formulation that constructs a temporal-level transport plan across consecutive frames and supervises both pairwise identity correspondences and its marginals.} It outperforms MVOT-TBD, a direct tracking-by-detection extension of M-MVOT.

\item Experiments demonstrate that M-MVOT and MVOT-Track outperform prior density map-based methods in localization and tracking, respectively, highlighting the benefits of precise localization and association.

\end{itemize}

\textit{A preliminary conference version of our work was presented at ECCV 2024}~\cite{MVOT24}.
\red{Compared with the conference version, this journal article includes the following new contributions:
(1) expanded analyses of M-MVOT under different loss settings, backbones, evaluation thresholds, and computational costs, together with comparisons against more recent methods;
(2) MVOT-TBD, which extends M-MVOT from localization to tracking by integrating ReID learning and identity association across frames;
(3) MVOT-Track, a tracking-specific OT loss that constructs a temporal transport plan between motion-warped predictions and ground-truth positions in the previous frame and directly supervises both the plan and its marginals; and
(4) comprehensive ablation studies of MVOT-Track covering the number of frames, view robustness, hyperparameter settings, and different tracking losses.}

The remainder of this paper is organized as follows. Section \ref{sec:method} presents the proposed multi-view optimal transport loss (M-MVOT) for crowd localization. Section \ref{sec:track} extends M-MVOT to the crowd tracking task and further introduces a tracking-specific OT loss, MVOT-Track. Section \ref{sec:experiments} provides comprehensive experimental analyses of the proposed localization and tracking methods, demonstrating their effectiveness. Finally, Section \ref{sec:conclusion} concludes the paper.

\section{Related Work}
\label{sec:relatedwork}


\subsection{Multi-view crowd localization}


\textbf{Traditional methods.}
Early traditional multi-view crowd localization methods first detect each person in the images and then associate the camera view results via correspondence matching with hand-crafted features \cite{viola2004robust, sabzmeydani2007detecting, fleuret2007multicamera, wu2007detection}.
W. Ge et al. \cite{Ge2010Crowd} presented a Bayesian detection and counting approach by sampling from a posterior distribution over crowd configurations.
Traditional methods often rely on hand-crafted features and background subtraction preprocessing, which limit their performance and application scenarios under severe occlusions.

\textbf{Deep learning methods.}
With the development of deep learning frameworks and more multi-view crowd datasets, deep learning-based multi-view crowd localization methods \cite{baque2017deep,chavdarova2017deep,song2021stacked} have seen great progress. 
The early deep-learning methods did not consider the feature alignment between camera views \cite{chavdarova2017deep}. 
MVMS \cite{zhang2019wide} proposed an end-to-end multi-view fusion framework with a feature projection layer for the wide-area crowd-counting task.
MVDet \cite{hou2020multiview} also used feature perspective transformations to fuse multi-view information for the multi-view crowd localization.
These are the first methods utilizing end-to-end frameworks for multi-view crowd localization tasks, which include single-view feature extraction, projection, and multi-view fusion and decoding.
In more recent work, SHOT \cite{song2021stacked} utilized multi-height projection with an attention-based soft selection module and replaced the perspective transformation with a homography transformation layer in the model for better localization performance.
MVDeTr  \cite{hou2021multiview} proposed the multi-view deformable transformer network MVDeTr for multi-view fusion by extending the deformable transformer framework in \cite{zhu2020deformable}. 
CVCS \cite{zhang2021cross} has studied multi-view crowd counting methods under the cross-view cross-scene setting, and proposed a large synthetic multi-view crowd dataset CVCS, which could also be used for multi-view crowd localization tasks.
3DROM \cite{qiu20223d} proposed a data augmentation method to relieve the model overfitting issue by generating random 3D cylinder occlusion masks in the 3D space.
Instead of projecting camera view representations onto a single \cite{hou2020multiview} or multiple \cite{song2021stacked} ground planes, TrackTacular \cite{Teepe_2024_CVPR} directly projects them into 3D space to obtain richer representations in the world coordinate system. Prior works generally extract multi-scale features in image space before projecting them onto the ground plane, whereas MSMVD \cite{yamane2025msmvd} explored multi-scale feature fusion directly in the ground plane space, which is neglected in earlier approaches. Since accurate camera calibration is challenging to acquire in practical settings, CaMuViD \cite{daryani2025camuvid} introduced a calibration-free multi-view localization framework, utilizing a shared 3D space to enable cross-view association in camera view space independent of camera parameters.



Overall, the existing multi-view crowd localization methods \cite{hou2020multiview, hou2021multiview, song2021stacked, qiu20223d} 
are trained with fixed-size Gaussian density maps as supervision.
However, using Gaussian density maps causes issues for localization in dense crowd regions, since when the Gaussian kernels significantly overlap, the local peaks will be smoothed out and the locations will be unrecoverable. Note that reducing the Gaussian width could potentially solve this problem, but it leads to a more difficult learning problem.
Considering that point-based supervision, such as optimal transport loss, achieved better performance
in the single-image crowd localization task, in this paper, we explore the multi-view optimal transport loss for the multi-view crowd localization task. 
\subsection{Single-image crowd localization}

Single-image crowd localization aims to estimate head locations using point annotations. 
Most existing methods are built upon Gaussian density map supervision 
\cite{zhang2015cross, sam2017switching, onoro2016towards, li2018csrnet, sindagi2017generating, shi2019revisiting, yan2019perspective}, 
whose predictions can be post-processed for localization. 
However, the intrinsic smoothness of Gaussian density maps often blurs local peaks in crowded regions, limiting localization accuracy.

Thus, point supervision methods \cite{ma2019bayesian, abousamra2021localization, song2021rethinking,liu2021bipartite, yan2022multiview, liu2023point}, blob segmentation \cite{laradji2018blobs} and inverse distance maps \cite{liang2022focal} have been proposed, many of which show better localization performance. 
Among them, optimal transport (OT)-based formulations~\cite{wang2020distribution, ma2021learning, wan2021generalized} have shown strong performance by directly minimizing the transport distance between predicted density maps and ground-truth point distributions. 
In particular, unbalanced OT (UOT) has been introduced to better handle count mismatches and perspective effects~\cite{ma2021learning, wan2021generalized}. 
Recent work such as OT-M ~\cite{Lin_2023_CVPR} also applies OT minimization to refine localization results from predicted density maps, formulating it as an iterative procedure that identifies the set of locations with minimal OT loss relative to the density map.

Despite OT's success in single-view crowd localization, point supervision with optimal transport for multi-view crowd localization is an unexplored area. The multi-view setting introduces new difficulties, such as variations in the camera view, projection errors due to poor camera calibration, and artifacts introduced when projecting features from the camera view to the 2D ground plane, which could hinder model training and localization ability.  

\textit{To address these issues, our proposed loss considers the camera view directions and distances when computing the transport cost, which is implemented via a Mahalanobis distance function.}

\subsection{Multi-view crowd tracking}

Existing multi-view crowd tracking methods can be divided into camera-view-first approaches and ground-plane tracking approaches.

\textbf{Camera-view-first tracking.}
A common paradigm first performs detection and tracking independently in each camera view, followed by cross-camera association. Representative works extend single-camera tracking frameworks with cross-view constraints or ID embedding modeling.
For example, MCTR~\cite{niculescu2025mctr} adopts a transformer-based multi-camera tracking architecture built upon DETR~\cite{carion2020end} with tracking embeddings, while other methods incorporate spatial-temporal constraints or human pose cues to improve cross-view consistency~\cite{cherdchusakulchai2024online, Kim_2023_CVPR, Kim_2024_CVPR}. 
Trajectory prediction has also been introduced to assist association across frames~\cite{Jeon_2023_CVPR}, and additional modalities such as infrared have been explored to improve robustness under low-light conditions~\cite{fan2025all}. 
Although effective, these above methods rely on accurate per-view tracking and subsequent cross-camera matching, making them sensitive to occlusion and view inconsistency in crowded scenes.

\textbf{Ground-plane tracking.}
To alleviate cross-view association ambiguity, another line of work performs tracking directly on the ground plane.
MVFlow~\cite{engilberge2023multi} models motion offsets on the ground plane to infer people’s presence, improving robustness to occlusions. However, it assumes that each person moves at most one grid cell per time step, which breaks down in realistic low-frame-rate scenarios where larger displacements occur.
\red{EarlyBird~\cite{Teepe_2024_WACV} projects multi-view image features into the world space for pedestrian localization and introduces a ReID branch for trajectory association. Its extension, TrackTacular~\cite{Teepe_2024_CVPR}, further lifts the ground-plane representation into 3D and predicts motion offsets to facilitate association.}
MVTrajecter \cite{Yamane_2025_ICCV} extends pairwise similarity computation in tracking to multiple temporal intervals by computing and fusing motion and appearance costs across different time steps. While this enhances robustness, it leads to increased computational overhead due to the need for evaluating multiple temporal pairs.

\textit{Compared with existing ground-plane multi-view crowd tracking methods, MVOT-Track formulates inter-frame identity association as an OT problem and explicitly supervises the temporal transport plan and its marginals.}

\section{Multi-view Optimal Transport}
\label{sec:method}

\begin{figure*}[t]
\begin{center}
   \includegraphics[width=0.85\linewidth]{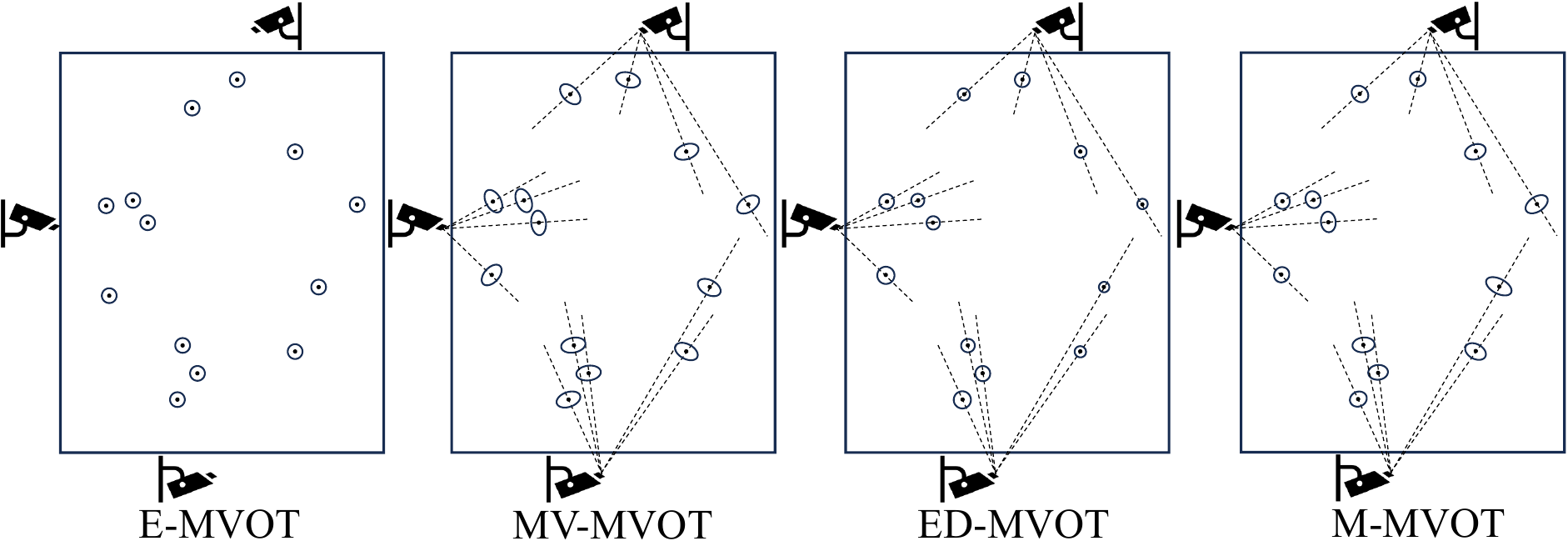}
\end{center}
   \caption{
   The comparison of different versions for the multi-view optimal transport, where the cost for each ground-truth point is calculated using the closest camera. 
   E-MVOT considers each point equally regardless of the camera views, which is the same as E-OT for a single camera view.
   MV-MVOT replaces the Euclidean distance with the Mahalanobis distance, and the direction is guided by the view ray direction. 
   ED-MVOT introduces the object-to-camera influence in the optimal transport cost of E-OT.
   M-MVOT considers both the view ray direction guidance and the object-to-camera distance adjustment in the optimal transport based on M-OT.
   }
\label{fig:mvot}
\end{figure*}

In this section, we first review optimal transport based on Euclidean distance cost (denoted as E-OT) for single-image crowd localization. Then, we introduce the proposed view ray Mahalanobis distance optimal transport (M-OT), where the cost can be guided either by the view ray direction of each GT location (MV-OT), or by the object-to-camera distance of each location (ED-OT), or both (M-OT). Finally, the proposed M-OT is applied under multi-views, where each prediction point's optimal transport cost is calculated under the closest camera view via M-OT, denoted as M-MVOT.

\subsection{Euclidean distance optimal transport}

We follow the loss formulation in \cite{wan2021generalized}, which is based on an unbalanced optimal transport formulation with Euclidean distance cost (E-OT). Denote ${\cal X} = \{(a_i, \textbf{x}_i)\}_{i=1}^n$ as the predicted density map, where $a_i$ is pixel $\textbf{x}_i$'s predicted density, $\textbf{a}=[a_i]_i$ is the density of all pixels, and $n$ is the number of pixels. We denote ${\cal Y} = \{(b_j, \textbf{y}_j)\}_{j=1}^m$ as the ground-truth dot map, where $b_j\in\{0,1\}$ indicates there is a person at location $\textbf{y}_j$, $\textbf{b}=[b_j]_j=\textbf{1}_m$, and $m$ is the ground-truth number of people. 
The entropic-regularized unbalanced optimal transport loss is as follows:
\begin{equation}
\begin{aligned}
  {\cal L}_{\bf{C}}^{\tau} = &\min_{\mathbf{P}^{loc} \in {\cal R}_+^{n\times m}} 
  \langle \textbf{C}, \mathbf{P}^{loc} \rangle 
  - \epsilon H(\mathbf{P}^{loc}) \\
     & + \tau \|\mathbf{P}^{loc}\bf{1}_m-\textbf{a}\|_2^2 \\
     & + \tau \|(\mathbf{P}^{loc})^T\bf{1}_n-\textbf{b}\|_1,
    \label{eqn:eot}
\end{aligned}
\end{equation}
where $\mathbf{P}^{loc} \in {\cal R}_+^{n\times m} $ is the transport plan matrix, which assigns each density $\textbf{x}_i$ of $\cal X$ to location $\textbf{y}_i$ of $\cal Y$ for measuring the cost. 
$\bf{C} \in {\cal R}_+^{n\times m} $ is the transport cost matrix, where $\textit{C}_{ij}$ measures the cost of moving the predicted density at $\textbf{x}_i$ to ground-truth dot location $\textbf{y}_j$. The cost function is defined as the exponential Euclidean distance between the predicted density of $\textbf{x}_i$ and the ground-truth dot $\textbf{y}_j$:
\begin{align}
C_{ij}=c(\textbf{x}_i, \textbf{y}_j)=\exp(\|\textbf{x}_i-\textbf{y}_j\|).
  \label{eqn:ed}
\end{align}
The second term $H(\mathbf{P}^{loc})=-\sum_{ij} P^{loc}_{ij}\log P^{loc}_{ij} $ is the entropic regularization term.
\cite{wan2021generalized} sets 
$D_1(\mathbf{P}^{loc}\bf{1}_m | \textbf{a})=\|\mathbf{P}^\mathit{loc}\bf{1}_m-\textbf{a}\|_2^2$, encouraging all predicted density values to have a corresponding annotation, and 
$D_2((\mathbf{P}^{loc})^T\bf{1}_n | \textbf{b})=\|(\mathbf{P}^\mathit{loc})^T\bf{1}_n-\textbf{b}\|_1$, encouraging all ground-truth points to be used in the transport plan. $\epsilon$ and $\tau$ are weighted factors to adjust the weight of the last three terms. The optimization of (\ref{eqn:eot}) can be computed via differentiable Sinkhorn updates \cite{peyre2017computational}, enabling efficient computation and end-to-end training.



\subsection{Mahalanobis distance optimal transport} \label{Sec:MD-OT}
The Euclidean distance in the transport cost matrix $\textbf{C}$ in (\ref{eqn:ed}) considers all directions of deviation from the ground-truth point equally. For multi-view crowd localization, we aim to apply the OT loss on the predicted density map on the ground plane. 
However, the projection step, which projects features from the camera view to the ground plane, introduces streaking artifacts in the feature map due to missing information, \textit{i.e.}, the unknown height of the corresponding object in the 3D world, which may lead to streaking artifacts in the density map that hinder its localization ability.
To counteract this effect, we use Mahalanobis distance to replace the Euclidean distance in (\ref{eqn:ed}), where the contours of the Mahalanobis distance are guided by the view ray direction at the location, denoted as MV-OT.
Besides, the object-to-camera distance could also adjust the calculation of the cost matrix for different regions, denoted as ED-OT. 
Finally, both the view ray directions and object-to-camera distances can be considered in the Mahalanobis distance optimal transport loss,
denoted as M-OT.

\textbf{View ray-directed transport cost (MV-OT).}
Using the Mahalanobis distance defines elliptical iso-contours in the transport cost, where certain directions will incur larger costs than others. 
Here, we set the ellipse shape to counteract the errors and streaking artifacts introduced by the projection step. 
Specifically, the short axis of the ellipse is set to be aligned with the view ray between the ground-truth point and the camera, and the long axis is set to be perpendicular to the view ray (MV-OT). In this way, the predictions that deviate from the ground-truth along the view ray from the ground-truth points are assigned with larger penalties (see ``MV-OT'' in Fig. \ref{fig:main_idea}, where more penalties are given along the view ray), in order to penalize prediction errors caused by projection distortion. 
Concretely, we first calculate the covariance matrix for each GT point:
\begin{equation}
\begin{aligned}
\mathbf{S} &= \mathbf{R}\mathbf{\Sigma}\mathbf{R}^{-1}, \\
\mathbf{\Sigma} &=
\begin{bmatrix}
\sigma_1^2 & 0 \\
0 & \sigma_2^2
\end{bmatrix}, \\
\mathbf{R} &=
\begin{bmatrix}
\cos\beta & -\sin\beta \\
\sin\beta & \cos\beta
\end{bmatrix},
\end{aligned}
\label{eqn:md}
\end{equation}
where $\bf{R} $ represents the rotation matrix from the original coordinate to the view ray coordinate, and $\beta$ is the rotation angle. $\bf{\Sigma}$ is the variance matrix with diagonal entries $\sigma_1^2$ and $\sigma_2^2$ corresponding to the variances along the view ray direction and perpendicular to the view ray direction, respectively. Here we set $\sigma_1^2 <\sigma_2^2$, and thus $1/\sigma_1^{2} > 1/\sigma_2^{2}$, indicating deviating predictions along the view ray direction incur the larger cost and thus are more heavily penalized.
Then, the Mahalanobis distance cost in MV-OT can be written as:
\begin{align}
C_{ij}=\exp(\sqrt{(\textbf{x}_i-\textbf{y}_j)^{\text{T}}\textbf{S}^{-1}(\textbf{x}_i-\textbf{y}_j)}).
  \label{eqn:md_cost_dis}
\end{align}
Note that when $\sigma_1^2 = \sigma_2^2=1$, the Mahalanobis distance degenerates into the Euclidean distance.
We choose $\sigma_1^2 = 1$ and $ \sigma_2^2 = 1.2$ in the experiments for MV-OT.

\textbf{Distance-adjusted transport cost (ED-OT).}
The object-to-camera distance also influences the localization accuracy for the points, where faraway points tend to have larger prediction errors. Thus, we use the object-to-camera distance to adjust the cost calculation and give more penalties to faraway regions. We rewrite $\sigma_1^2$ and $\sigma_2^2$ as follows. 
\begin{equation}
\sigma_1^2 = \sigma_2^2 = 1 / \exp(\alpha*\MinMaxNorm(d_{cam})),
\label{eqn:md_dis}
\end{equation} 
where $d_{cam}$ denotes the distance from the ground-truth point to the camera, which can be obtained from the camera calibrations and point coordinates, and $\MinMaxNorm$ normalizes the object-to-camera distance into $[0, 1]$ through ($d_{cam}$ - $\min(d_{cam})$) / ($\max(d_{cam})$ - $\min(d_{cam})$). Here $\alpha$ is an adaptive factor to adjust the influence of distance to the cost function in (\ref{eqn:md_dis}) and (\ref{eqn:md_dir_dis}). When $d_{cam}$ is larger, $\sigma_1^2$ and $\sigma_2^2$ decrease, and thus more penalties are assigned. Since $\sigma_1^2 = \sigma_2^2$, the Euclidean distance is actually used in the cost, but the object-to-camera distance adjusts the Euclidean distance for different points, which we denote as ED-OT.

\textbf{View ray-directed and distance-adjusted transport cost (M-OT).}
To consider both the influence of the view ray direction and object-to-camera distance, we use the object-to-camera distance to adjust the long-axis and short-axis extents of the ellipse iso-contours in the Mahalanobis distance. Specifically, for the GT points that are far away from the camera ($d_{cam}$ is larger), a larger variance is assigned perpendicular to the view ray ($\sigma_2^2$), which has the effect of penalizing deviations perpendicular to the view ray less, and penalizing deviations along the view ray ($\sigma_1^2$) more: 
\begin{equation}
\scalebox{0.9}{$
\sigma_1^2  = 1, 
\quad
\sigma_2^2 = \exp(\alpha*\MinMaxNorm(d_{cam})).$}
\label{eqn:md_dir_dis}
\end{equation}  
By combining (\ref{eqn:md}) and (\ref{eqn:md_dir_dis}), we obtain the view ray-directed and distance-adjusted Mahalanobis distance cost function for OT, denoted as M-OT. 
Note that when $\alpha=0$, then  $\sigma_1^2=\sigma_2^2=1$, and the proposed Mahalanobis distance cost of M-OT degenerates into E-OT in (\ref{eqn:ed}).




\subsection{Mahalanobis distance multi-view optimal transport (M-MVOT)} 
We introduce the Mahalanobis distance optimal transport (M-OT) for single views in Sec.~\ref{Sec:MD-OT}.
We further introduce the Mahalanobis distance multi-view optimal transport (M-MVOT) by combining the proposed Mahalanobis distance optimal transport via a distance-based selection strategy.
Specifically, the cost of a point is calculated using the closest camera view's M-OT defined as follows.
\begin{equation}
\scalebox{0.8}{$
C_{ij} = \sum_{k=1}^{K} \mathbbm{1}(d^k_{\text{cam}}) 
\exp\Big(\sqrt{(\mathbf{x}_i - \mathbf{y}_j)^{\top} \mathbf{S}_k^{-1} (\mathbf{x}_i - \mathbf{y}_j)}\Big),$}
\label{eqn:dist_select}
\end{equation}
where the binary indicator $\mathbbm{1}(d^k_{\text{cam}})$ selects the closest camera:
\begin{equation}
\scalebox{0.9}{$
\mathbbm{1}(d^k_{\text{cam}}) = 
\begin{cases}
1, & \text{if } d^k_{\text{cam}} = \min_{p \in \{1,\dots,K\}} d^p_{\text{cam}},\\
0, & \text{otherwise}.
\end{cases}$}
\end{equation}
Here, $K$ is the number of cameras, and $d^k_{\text{cam}}$ denotes the distance from the ground-truth point to camera $k$.

Following the strategy, we extend E-OT, MV-OT, ED-OT, and M-OT to multi-camera view settings for multi-view crowd localization tasks, denoted as E-MVOT, MV-MVOT, ED-MVOT, and M-MVOT, respectively, where the closest camera is used to compute the cost function for a given ground-truth point, as in Fig. \ref{fig:mvot}.

\section{MVOT-based Multi-view Crowd Tracking}
\label{sec:track}

\red{In this section, we extend the M-MVOT-based localization framework to multi-view crowd tracking through two methods. MVOT-TBD is a tracking-by-detection baseline that combines the M-MVOT-supervised localization model with a ReID module and a two-stage association algorithm following EarlyBird~\cite{Teepe_2024_WACV}. MVOT-Track further formulates inter-frame identity association as an optimal transport problem and explicitly supervises temporal correspondences across adjacent frames.}

\begin{figure}[t]
\begin{center}
   \includegraphics[width=0.95\linewidth]{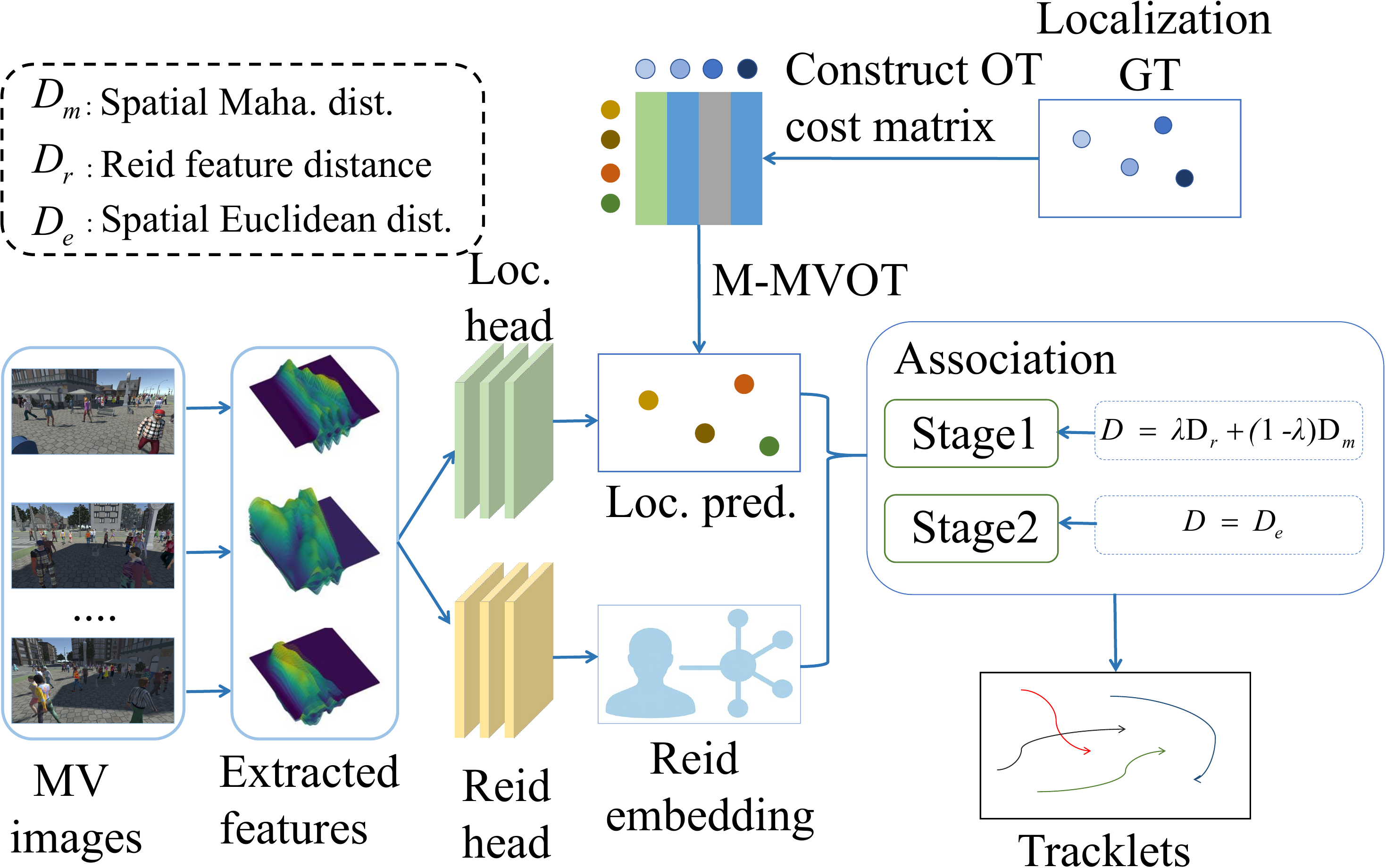}
\end{center}
   \caption{The overall architecture of MVOT-TBD for multi-view crowd tracking.
MVOT-TBD integrates a ReID head into the localization framework and leverages the extracted ReID embeddings to perform two-stage association for tracking.
The localization head is trained using our proposed M-MVOT loss.}
\label{fig:mvot_tbd}
\end{figure}

\begin{figure*}[t]
\begin{center}
   \includegraphics[width=1.0\linewidth]{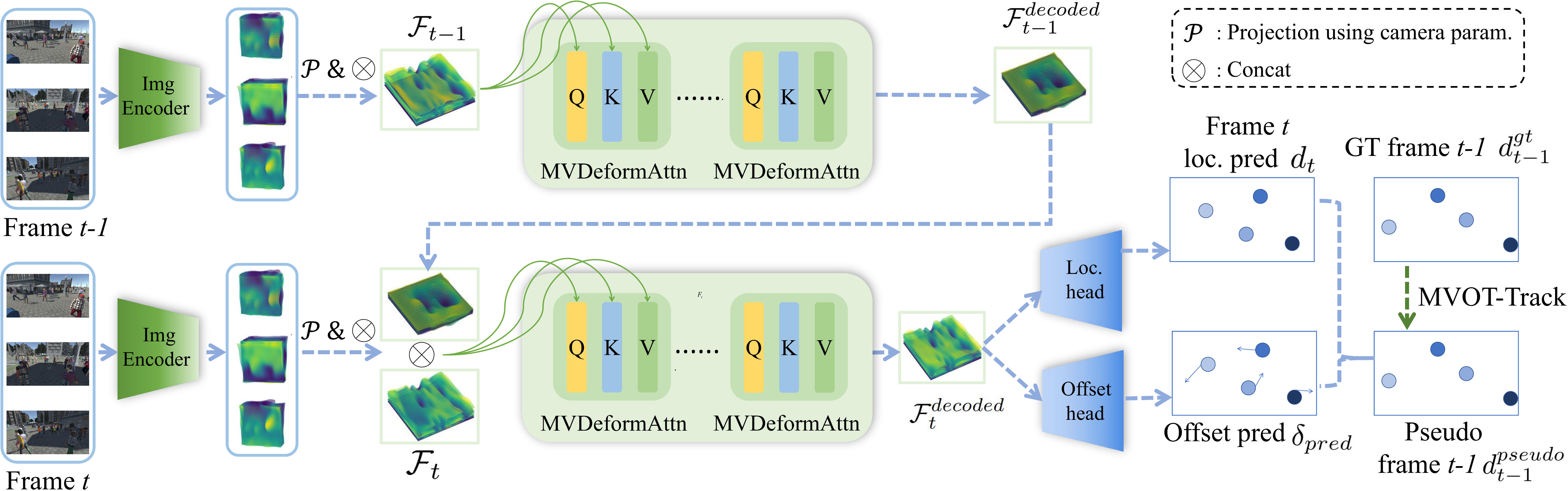}
\end{center}
   \caption{The overall multi-view crowd tracking framework with MVOT-Track loss is as follows. The model takes two adjacent frames, $t-1$ and $t$, as input. Both frames are processed by view feature extraction and ground-plane projection, following the same pipeline as multi-view localization, resulting in $\mathcal{F}_{t-1}$ and $\mathcal{F}_{t}$. Next, $\mathcal{F}_{t-1}$ is fed into MVDeformAttn to enable cross-view feature interaction, producing $\mathcal{F}_{t-1}^{decoded}$. This feature is then concatenated with $\mathcal{F}_{t}$ and further processed by MVDeformAttn layers for temporal fusion, yielding temporally enriched features $\mathcal{F}_{t}^{decoded}$.
  The fused features are passed into two prediction heads: one for crowd localization at frame $t$, and the other for estimating motion offsets toward frame $t-1$. The predicted motion offsets are used to warp the localization results back to frame $t-1$. Finally, the MVOT-Track loss minimizes the optimal transport distance between the warped pseudo predictions and the ground truth at frame $t-1$, with additional regularization applied to the transport matrix $\mathbf{P}^{tra}$.
}

\label{fig:mvot_track_pipeline}
\end{figure*}

\subsection{MVOT-TBD}
\red{As shown in Fig.~\ref{fig:mvot_tbd}, MVOT-TBD extends the M-MVOT-based localization framework with a person ReID head and a two-stage association algorithm to link localized pedestrians across consecutive frames.}

The ReID head aims to extract discriminative identity features for each tracked individual. 
To obtain reliable ReID features that can be sampled at ground-plane locations during inference, 
we jointly train the ReID head with the entire model. 
Specifically, we adopt two supervision strategies: a classification loss and a supervised contrastive loss. 
For each detected instance $i$ in a training batch of size $N$, we extract a ReID feature vector 
$\mathbf{f}_i \in \mathbb{R}^d$ at the corresponding ground-plane detection location, 
where $d$ denotes the feature dimension. 
The feature $\mathbf{f}_i$ is then fed into a linear classification layer to produce identity logits 
$\mathbf{s}_i \in \mathbb{R}^{N_{\text{id}}}$, where $N_{\text{id}}$ denotes the total number of unique identities, 
with each individual treated as an independent class in the classification space. 
We denote by $\mathbf{s}_{i,j}$ the logit corresponding to the $j$-th identity, and 
$y_i \in \{1, \dots, N_{\text{id}}\}$ the ground-truth identity label of instance $i$. 
The classification loss is formulated as the standard cross-entropy loss:
\begin{equation}
\mathcal{L}_{\text{cls}} = -\frac{1}{N}\sum_{i=1}^{N}
\log
\frac{\exp(\mathbf{s}_{i,y_i})}{\sum_{j=1}^{N_{\text{id}}}\exp(\mathbf{s}_{i,j})}.
\end{equation}
In addition, we apply a supervised contrastive loss~\cite{khosla2020supervised} on the decoder features, encouraging features of the same identity that are sampled from both the ground plane and the camera views to cluster closely, while separating those of different identities:
\begin{equation}
\scalebox{0.9}{$
\mathcal{L}_{\text{SupCon}} 
= \sum\limits_{i \in I} \frac{-1}{|\mathcal{P}_{\text{pos}}(i)|} 
\sum\limits_{p \in \mathcal{P}_{\text{pos}}(i)} 
\log \frac{\exp \left( \mathbf{f}_i^\top \mathbf{f}_p / \tau \right)}
{\sum\limits_{a \in A(i)} \exp \left( \mathbf{f}_i^\top \mathbf{f}_a / \tau \right)} \,,$}
\end{equation}
where $\mathcal{P}_{\text{pos}}(i)$ denotes the set of positive samples sharing the same identity with anchor $i$ (including both ground-plane and camera-view samples), $A(i)$ represents the set of all samples in the batch excluding the anchor $i$, $\tau$ is a temperature parameter, and $\mathbf{f}_i$ is the feature embedding of sample $i$.  
The overall ReID objective is a sum of the classification and contrastive losses as in \cite{Teepe_2024_WACV}.
We use a standard Kalman filter \cite{kalman1960new} here to estimate the position and velocity of an object from noisy observations,
where the bounding box coordinates are directly taken as observations of the object state. For each target, the state is modeled as
\begin{equation}
\mathbf{x} = [x, y, a, h, \dot{x}, \dot{y}, \dot{a}, \dot{h}]^{T},
\end{equation}
where $(x, y)$ denotes the center position of the bounding box, $a$ is the aspect ratio, $h$ is the height, and $(\dot{x}, \dot{y}, \dot{a}, \dot{h})$ represent the corresponding velocities. Although predictions are performed on ground-plane points, we assign a fixed bounding box size to each detection for state representation and tracking.

For frame association, we implement a two-stage association strategy using the Hungarian algorithm \cite{kuhn1955hungarian}. In the first stage, matching costs between frames are computed by combining appearance and motion cues. Appearance distance is defined as
\begin{align}
    D_r(\mathbf{f}_i, \mathbf{f}_j) = 1 - \frac{\mathbf{f}_i^\top \mathbf{f}_j}{\|\mathbf{f}_i\|\|\mathbf{f}_j\|}, 
\end{align}
where $\mathbf{f}_i$ and $\mathbf{f}_j$ are ReID features of a tracklet and a detection. Motion distance is
\begin{align}
D_m (\mathbf{x}_i, \mathbf{y}_j) = (\mathbf{d}_j - \mathbf{y}_i)^\top \mathbf{S}_i^{-1} (\mathbf{d}_j - \mathbf{y}_i),
\end{align}
where $(\mathbf{y}_i, \mathbf{S}_i)$ are the projected mean and covariance of the $i$-th tracklet estimated by Kalman filter, and $\mathbf{d}_j$ denotes the $j$-th detection. These two are fused as $D = \lambda D_r + (1-\lambda) D_m$, balancing appearance and motion, with gating applied before Hungarian matching. 
In the second stage, unmatched tracks and detections are associated spatially using the Euclidean distance between target centers with a relaxed spatial threshold, which helps handle appearance variations and improves robustness in challenging cases.

\subsection{MVOT-Track}

\red{MVOT-TBD performs localization followed by trajectory association and therefore does not directly optimize inter-frame correspondences. To address this limitation, we introduce MVOT-Track, a tracking-specific optimal transport formulation that models identity association between adjacent frames. Unlike M-MVOT, whose transport plan matches predicted density with ground-truth points within a single frame, MVOT-Track constructs a temporal transport plan between motion-warped predictions at frame $t$ and ground-truth positions at frame $t-1$, and directly supervises the plan and its marginals to capture target-presence changes across frames.}

\textbf{Model architecture.} 
To enable the use of this loss, we extend the existing backbone MVDeTr \cite{hou2021multiview} to the temporal domain, as illustrated in Fig.~\ref{fig:mvot_track_pipeline}. MVDeTr is originally designed for multi-view localization, where a ground-plane decoder composed of several multi-view deformable attention layers is employed.
In contrast to the localization setting, which processes a single frame at a time, our approach takes two consecutive frames as input to capture temporal information, denoted as $I_{t-1}$ and $I_t$, corresponding to the previous and current frames, respectively. These frames are first encoded by an image encoder and then projected onto the ground plane, following the same procedure as in multi-view localization. Subsequently, the ground-plane features from the two frames are jointly processed by multi-view deformable attention layers to perform temporal fusion.

\begin{figure*}[t]
\begin{center}
   \includegraphics[width=1\linewidth]{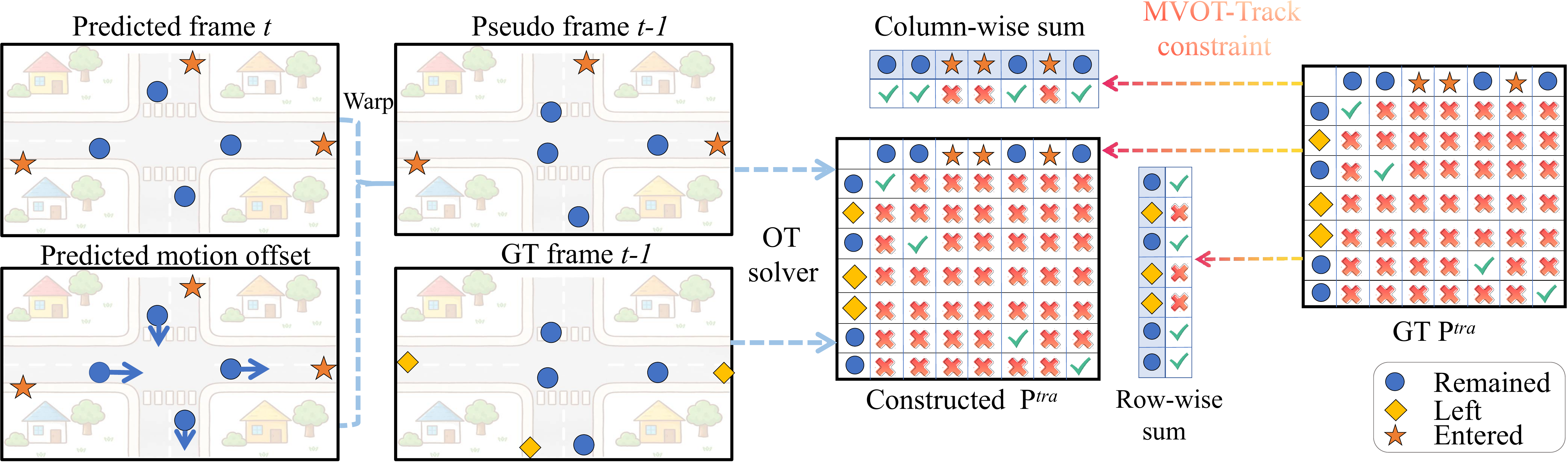}
\end{center}
   \caption{\textbf{The illustration of temporal correspondence in tracking.} The predicted result at frame $t$ is warped to frame $t{-}1$ using the predicted motion offsets. We then compute the correspondence between the warped frame $t{-}1$ and the ground-truth annotations at frame $t{-}1$. The blue circles indicate individuals who remain in the scene across both frames. The yellow diamonds denote individuals who exit the scene at frame $t{-}1$, while the orange stars represent individuals who newly appear at frame $t$. The ground-truth correspondence matrix is a binary matrix, where an entry of \ding{51} indicates a valid identity match across frames, and \ding{55} indicates no correspondence.}
\label{fig:mvot_track}
\end{figure*}

\textbf{Multi-view deformable attention.} 
We follow the formulation of the multi-view deformable attention defined in MVDeTr \cite{hou2021multiview}:
\small{
\begin{align}
\begin{split}
&\mathrm{MVDeformAttn}(f_{\hat{c}}, p, \hat{c}) \\
&\quad = \sum_{m=1}^{M} W_m \sum_{c'=1}^{C} \sum_{k=1}^{K} 
a_{mkc'} W'_m f^{c'}(p + \Delta p_{mkc'}), \\
&\quad \text{subject to: } \sum_{k=1}^{K} \sum_{c'=1}^{C} a_{mkc'} = 1,
\end{split}
\label{eqn:mvdeformattn}
\end{align}}
where $f_{\hat{c}}$ denotes the fused ground-plane feature at query position $p$ for the current view $\hat{c}$, and $f^{c'}$ is the projected ground-plane feature from source camera $c'$. 
For each attention head $m \in \{1,\dots,M\}$, $K$ reference points $\Delta p_{mkc'}$ are sampled from each camera view $c'$ and projected via $W'_m$. 
The attention weights $a_{mkc'}$ indicate the relative importance of each reference point and satisfy $\sum_{k=1}^{K}\sum_{c'=1}^{C} a_{mkc'} = 1$, ensuring a soft selection across reference points and views. 
The weighted features are summed within each head, and outputs of all $M$ heads are aggregated via learned weight $W_m$. 
Elliptical distortions along camera view rays involved by ground-plane projection can be alleviated by dynamically offsetting the reference points $\Delta p_{mkc'}$.

\textbf{Temporal fusion of multi-view features.} 
Inspired by the Late-to-Early fusion strategy \cite{he2023lef, Teepe_2024_CVPR}, 
we perform temporal fusion on multi-view ground-plane features.
Let $\mathcal{F}_{t-1} = \{ f_{t-1}^{c'} \}_{{c'}=1}^{C}$ and 
$\mathcal{F}_{t} = \{ f_{t}^{{c'}} \}_{{c'}=1}^{C}$ denote the sets of projected 
ground-plane feature maps from all camera views at frames $t-1$ and $t$, respectively. 
We first apply $L$ stacked multi-view deformable attention layers to $\mathcal{F}_{t-1}$ to aggregate multi-view information:
\begin{equation} 
\begin{aligned}
\mathcal{F}_{t-1}^{decoded} 
&= \mathrm{MVDeformAttn}^{(L)} \Big( 
       \dots \\
&\quad \mathrm{MVDeformAttn}^{(1)}(\mathcal{F}_{t-1}) \dots \Big),
\end{aligned}
\end{equation}
where each MVDeformAttn layer operates jointly over all camera features. 
In our implementation, we set $L = 3$. 
The decoded features $\mathcal{F}_{t-1}^{decoded}$ are then fused with 
$\mathcal{F}_{t}$ via channel-wise concatenation, denoted by $\oplus$, 
and passed through the same MVDeformAttn layers again to obtain the final 
decoded representation:
\begin{equation} 
\scalebox{1}{
$ \begin{aligned}
\mathcal{F}_{t}^{decoded} 
&= \mathrm{MVDeformAttn}^{(L)} \Big( 
       \dots \\
&\mathrm{MVDeformAttn}^{(1)}(\mathcal{F}_{t-1}^{decoded} \oplus \mathcal{F}_{t}) \\
\dots \Big),
\end{aligned} $
}
\end{equation}
where $\oplus$ represents concatenation along the channel dimension. 
The resulting $\mathcal{F}_{t}^{decoded}$ encodes both multi-view and 
temporal information from adjacent frames. It is subsequently fed into 
two task-specific heads: one for predicting the ground-plane density map 
for crowd localization, and the other for estimating per-person motion 
offsets from frame $t$ to frame $t\!-\!1$, enabling crowd association 
for tracking.

\begin{table*}
\setlength{\tabcolsep}{2pt}
\small
\centering
\caption{Overview of the proposed localization and tracking variants. 
Localization methods differ in the variance defined in the covariance matrix and whether distance-based camera selection is applied. 
Tracking variants extend the framework to temporal association modeling.}
\begin{tabular}{l|lc}
\hline
\multicolumn{3}{c}{\textbf{Localization Variants}} \\
\hline
Method & Variance & Cam Selection \\
\hline
E-OT      & $\sigma_1^2 = \sigma_2^2 = 1$ & without \\
MV-OT     & $\sigma_1^2=1, \sigma_2^2 = 1.2$ & without \\
ED-OT     & $\sigma_1^2 = \sigma_2^2 = 1 / \exp(\alpha*\MinMaxNorm(d_{cam}))$ & without \\
M-OT      & $\sigma_1^2 = 1, \sigma_2^2 = \exp(\alpha*\MinMaxNorm(d_{cam}))$ & without \\
\hline
E-MVOT    & $\sigma_1^2 = \sigma_2^2 = 1$ & with \\
MV-MVOT   & $\sigma_1^2=1, \sigma_2^2 = 1.2$ & with \\
ED-MVOT   & $\sigma_1^2 = \sigma_2^2 = 1 / \exp(\alpha*\MinMaxNorm(d_{cam}))$ & with \\
M-MVOT    & $\sigma_1^2 = 1, \sigma_2^2 = \exp(\alpha*\MinMaxNorm(d_{cam}))$ & with \\
\hline
\multicolumn{3}{c}{\textbf{Tracking Variants}} \\
\hline
MVOT-TBD   & \multicolumn{2}{l}{\red{Tracking-by-detection framework based on M-MVOT with ReID and two-stage association}} \\
MVOT-Track & \multicolumn{2}{l}{\red{Tracking-specific OT loss supervising the temporal transport plan and its marginals}} \\
\hline
\end{tabular}
\label{table:brief}
\end{table*}

\textbf{Loss definition.} To optimize the above architecture, we first construct a pseudo localization prediction at frame $t\!-\!1$ by warping the current frame $t$ localization predictions using the predicted motion offsets, which means $d_{t-1}^{pseudo} = d_{t} + \delta_{pred} $. $\delta$ represents the offsets. We then compute the optimal transport loss between the constructed pseudo localization $d_{t-1}^{pseudo}$ and the ground-truth $d_{t-1}^{gt}$. It is crucial to recognize that the optimal transport between the pseudo localization at frame $t\!-\!1$ and the ground-truth localization at frame $t\!-\!1$ is inherently unbalanced. This is because some individuals in frame $t\!-\!1$ may have exited the scene and thus have no corresponding matches in frame $t$, while frame $t$ may include newly arrived individuals who have no counterparts in the previous frame (Fig. \ref{fig:mvot_track}).

This phenomenon is specific to tracking and does not arise in localization tasks, which neither model per-person identity nor enforce temporal consistency. To address this, we formulate the people temporal association problem as an optimal transport task and introduce three dedicated regularization terms. The formulation of MVOT-Track is as follows:
\begin{equation}
\scalebox{0.9}{
$\begin{aligned}
  {\cal L}_{\bf{C}}^{\tau}(d_{t-1}^{pseudo}, d_{t-1}^{gt}) = 
  &\min_{\textbf{P}^{tra} \in {\cal R}_+^{n\times m}} 
    \langle \textbf{C}, \textbf{P}^{tra} \rangle -  \epsilon H(\textbf{P}^{tra}) \\
     & + \gamma \|\textbf{P}^{tra} \mathbf{1}_m - \textbf{P}_{GT}^{tra} \mathbf{1}_m \|_1 \\
     & + \gamma \|(\textbf{P}^{tra})^T \mathbf{1}_n - (\textbf{P}_{GT}^{tra})^T \mathbf{1}_n \|_1 \\
     & + \psi \|\textbf{P}^{tra} - \textbf{P}_{GT}^{tra}\|_1
\end{aligned}$
}
\label{mvot_track_formula}
\end{equation}
$\textbf{C}$ is defined as the Euclidean distance between $d_{t-1}^{\text{pseudo}}$ and $d_{t-1}^{\text{gt}}$. Here, $\textbf{P}^{tra}$ \red{is the temporal transport plan, which represents the correspondence between adjacent frames.} Note that we do not compute the correspondence directly between the original localization results of adjacent frames. Instead, we first warp the current frame's prediction to the previous frame using the predicted offsets, and then compute the correspondence. This design encourages the offsets prediction to be more accurate by explicitly aligning the warped predictions with the ground truth.

We introduce three additional regularization terms on $\textbf{P}^{tra}$. The first two terms supervise the marginals $\textbf{P}^{tra}\mathbf{1}_m$ and $(\textbf{P}^{tra})^T\mathbf{1}_n$, which are $n$- and $m$-dimensional vectors respectively. Here, $\textbf{P}^{tra}\mathbf{1}_m$ represents the retention status of individuals from frame $t\!-\!1$: each value equals to 1 indicates the person remains at frame $t$, while values equal to 0 imply the person has exited the scene. Similarly, $(\textbf{P}^{tra})^T\mathbf{1}_n$ indicates the origin of individuals in frame $t$: zero value suggest a new arrival not present in frame $t\!-\!1$. By supervising these marginals, the network is encouraged to distinguish between persistent individuals and those entering or exiting the scene. The third term directly supervises the transport plan $\textbf{P}^{tra}$ itself, encouraging accurate pairwise associations between individuals across frames based on the ground-truth correspondence matrix $\textbf{P}_{GT}^{tra}$. This matrix is constructed from identity-level annotations and reflects the true temporal correspondences. $\gamma$, and $\psi$ are weighting hyperparameters. These three terms work together to encourage the model to learn temporally consistent assignments.

The overall loss related to tracking can be formulated as:
\begin{equation}
\begin{aligned}  
{\cal L}_{tracking} = {\cal L}_{\bf{C}}^{\tau}(d_{t-1}^{pseudo}, d_{t-1}^{gt}) + {\cal L}_{offset}^{L_1},
\label{eqn:l_tracking}
\end{aligned}
\end{equation}
{where ${\cal L}_{offset}^{L_1}$ denotes the $L_1$ distance between the predicted and ground-truth offsets, which is commonly adopted in previous works \cite{Teepe_2024_CVPR, 10742444}. We combine it with the proposed MVOT-Track loss ${\cal L}_{\bf{C}}^{\tau}(d_{t-1}^{pseudo}, d_{t-1}^{gt})$ to jointly supervise the crowd correspondences across adjacent frames for tracking.}

\section{Experiments and Results}

\label{sec:experiments}

\subsection{Methods overview}

For better clarity, we summarize all variants of our framework in Table~\ref{table:brief}. 
The localization variants (E-OT, MV-OT, ED-OT, M-OT, E-MVOT, MV-MVOT, ED-MVOT, and M-MVOT) 
mainly differ in the variance defined in the covariance matrix when computing the transport cost matrix, 
as well as whether distance-based camera selection is applied for multi-camera fusion. 
Here, $\sigma_1^2$ denotes the variance along the camera view ray, and $\sigma_2^2$ denotes the variance perpendicular to the view ray.
In addition, we extend the framework to tracking with two variants, MVOT-TBD and MVOT-Track. 
These tracking variants focus on temporal association modeling rather than variance design or camera selection.

\subsection{Experiment settings}

\textbf{Datasets.}
We evaluate the proposed methods on the multi-view localization task on 3 public datasets in the experiments, including CVCS \cite{zhang2021cross}, Wildtrack \cite{chavdarova2018wildtrack}, and MultiviewX \cite{hou2020multiview}.
Wildtrack and MultiviewX are also used for evaluating the multi-view crowd tracking performance.
\textbf{CVCS} is a synthetic multi-view crowd dataset, containing 31 scenes, where 23 are for training and the remaining 8 for testing. Each scene contains 100 multi-view frames with about 60-120 camera views. The size of the scenes in CVCS varies from 10m$\times$20m to 90m$\times$80m. Each scene contains 90-180 people. 
\textbf{Wildtrack} is a real-world single-scene multi-view dataset, whose scene size is  12m$\times$36m. There are 20 people per frame on average in the dataset. The dataset consists of 7 camera views and 400 synchronized multi-view frames, in which the first 360 frames are used for training and the 40 frames are for testing. 
\textbf{MultiviewX} is a synthetic single-scene multi-view dataset, whose scene size is 16m$\times$25m and consists of 6 camera views.
The dataset includes 400 multi-view frames, and the first 360 are used for training and the last 40 are for testing. MultiviewX has around 40 synthetic people per frame. 
The resolution of the camera view images in the 3 datasets is 1920$\times$1080.
Note that CVCS is a \emph{cross-scene} dataset, where the same scene and camera layout do not appear in both the training set and test set. In contrast, both Wildtrack and MultiviewX are \emph{single-scene} datasets with the same camera layout for both training and testing. 


\textbf{Comparison methods.}
We first compare the proposed M-MVOT loss
based \textit{multi-view crowd localization} model against the Euclidean distance version (E-MVOT) using the same architecture, as well as against existing methods MVDet \cite{hou2020multiview}, SHOT \cite{song2021stacked}, MVDeTr \cite{hou2021multiview}, 3DROM \cite{qiu20223d}, EarlyBird \cite{Teepe_2024_WACV}, and TrackTacular \cite{Teepe_2024_CVPR}.
We report the performance of previous methods on Wildtrack and MultiviewX according to the published results, and also apply them on the CVCS dataset for comparison. 
We also compare with RCNN \cite{xu2016multi}, POM-CNN \cite{fleuret2007multicamera}, DeepMCD \cite{chavdarova2017deep}, DeepOcc.~\cite{baque2017deep}, and Volumetric \cite{iskakov2019learnable} on the Wildtrack and MultiviewX datasets. 

{For \textit{multi-view crowd tracking}, we compare MVOT-Track with MVOT-TBD and existing multi-view crowd tracking SOTAs, including KSP-DO \cite{chavdarova2018wildtrack}, KSP-DO-ptrack \cite{chavdarova2018wildtrack}, GLMB-YOLOv3 \cite{ong2020bayesian}, GLMB-DO \cite{ong2020bayesian}, DMCT \cite{you2020real}, DMCT Stack \cite{you2020real}, ReST \cite{Cheng_2023_ICCV}, EarlyBird \cite{Teepe_2024_WACV}, MVTr \cite{yang2024end}, MVFlow \cite{engilberge2023multi}, TrackTacular \cite{Teepe_2024_CVPR} and MVTrajecter \cite{Yamane_2025_ICCV}.}
\textit{Note that, unlike other comparison methods, MVTrajecter~\cite{Yamane_2025_ICCV} does not release its code or pretrained models. \red{Therefore, we report both the original and our reproduced results on Wildtrack, while some implementation details cannot be fully determined from the publication alone.}}

\textbf{Implementation details.}
In the experiments, for CVCS, the input image resolution is 1280$\times$720.
In the training, 5 views are randomly selected for 5 times in each iteration per frame of each scene, and the same number of views (5) is randomly selected for 21 times per frame in the evaluation.
For Wildtrack, the input image resolution is resized to $1280\!\times\!720$, and the scene map resolution is $ 120\!\times\!360$, where each pixel represents 0.1m in the real world.
For MultiviewX, the input image resolution is also resized to $1280\!\times\!720$, and the scene map resolution is $ 160\!\times\!250$, where each pixel is 0.1m in the real world.
The map distance threshold is 0.5m in the real world as in \cite{hou2020multiview} for all methods on the 3 datasets. On the CVCS dataset, the multi-view crowd localization model is implemented based on the existing multi-view crowd localization method EarlyBird \cite{Teepe_2024_WACV}.
On the Wildtrack and MultiviewX datasets, the implementation for the multi-view crowd localization is based on MVDeTr \cite{hou2021multiview}. 
The ground-plane density map prediction losses in \cite{hou2020multiview,hou2021multiview} are replaced with the proposed Mahalanobis distance multi-view optimal transport loss (M-MVOT).
The models are trained with the proposed M-MVOT loss and the auxiliary 2D density map losses in \cite{hou2020multiview} together.
For Wildtrack and CVCS, $\tau$ is set to 1, and $\tau$ is set to 20 for MultiviewX.
$\alpha$ is 1 on CVCS and 0.05 on both MultiviewX and Wildtrack. 
For multi-view crowd tracking, in MVOT-TBD, $\lambda$ is set to 0.98, and the relaxed threshold for the second-stage association is set to 2.5 m. For MVOT-Track, each batch randomly samples two consecutive frames in each epoch. We set $\gamma=10$ and $\psi=1$ for both MultiviewX and Wildtrack. All other settings are consistent with those used for multi-view crowd localization.


\begin{table*}[t]
\centering
\scriptsize
\caption{The multi-view crowd localization results on the CVCS dataset. The distance threshold is set to 0.5m in the world coordinate system for all methods. Our methods achieve the highest MODA and F1 score values.}

\begin{tabular}{l|ccccc}
\hline
    Method                                   & MODA$\uparrow$    & MODP$\uparrow$       & Precision$\uparrow$        & Recall$\uparrow$    & F1\_score$\uparrow$ \\
\hline
    MVDet \cite{hou2020multiview}                 & 14.2   & 59.3     & 85.0     & 17.3      & 28.7 \\
    SHOT \cite{song2021stacked}                    & 31.7   & 72.1     & 94.5     & 33.6      & 49.6  \\
    MVDeTr  \cite{hou2021multiview}               & 24.9    & \textbf{79.6}     & \textbf{98.1}     & 25.4      & 40.4  \\
    3DROM \cite{qiu20223d}                        & 20.1   & 74.2    & 84.1     & 23.7      & 37.0  \\
    EarlyBird \cite{Teepe_2024_WACV}     & \underline{44.8}    & \underline{77.2}     & 95.5      & 47.0   & 63.0   \\
    TrackTacular \cite{Teepe_2024_CVPR}     &42.1     &\underline{77.2}      &\underline{95.8}       &44.1    &  60.4\\
\hline
\hline
    \red{E-MVOT (ECCV24)}   &43.1 	&74.3      &85.6      &51.8      &64.5    \\
    E-MVOT (Ours)  &42.8 	&72.4      &76.7      &\textbf{61.4}      &\underline{68.2}    \\
    \red{M-MVOT (ECCV24)}   & 43.5	  & 74.1     & 85.5    & 52.3      & 64.9    \\
    M-MVOT (Ours)   & \textbf{45.8}	  & 73.2     & 82.0    & \underline{58.7}      & \textbf{68.4}    \\
\hline 
\end{tabular}
\label{table:CVCS_results}
\end{table*}

\begin{table*}[t]
\scriptsize
\centering
\setlength{\tabcolsep}{5pt}
\caption{Comparison of the multi-view crowd localization performance on single-scene datasets Wildtrack and MultiviewX. The proposed method is the best on MultiviewX and comparable on Wildtrack. The distance threshold is 0.5m in the real world \cite{hou2020multiview}. 
}
\begin{tabular}{l|ccccc|ccccc}

\hline
    Dataset &  \multicolumn{5}{c|}{MultiviewX}  &  \multicolumn{5}{c}{Wildtrack}  \\
    Method         & MODA$\uparrow$     & MODP$\uparrow$         & P.$\uparrow$         & R.$\uparrow$      & F1.$\uparrow$ 
                & MODA$\uparrow$     & MODP$\uparrow$         & P.$\uparrow$         & R.$\uparrow$      & F1.$\uparrow$  \\
\hline
    RCNN \cite{xu2016multi}                    & 18.7    & 46.4        & 63.5         &  43.9  & 51.9
                                              & 11.3    & 18.4        & 68         &  43  & 52.7
                                                   \\
    POM-CNN \cite{fleuret2007multicamera}    &  -  &    -      & -     & -  & -    &  23.2  &    30.5      & 75     & 55  & 63.5
                                                        \\
    DeepMCD  \cite{chavdarova2017deep}       & 70.0   & 73.0         & 85.7         & 83.3  & 84.5    & 67.8   & 64.2         & 85         & 82  & 83.5
                                                  \\
    DeepOcc. \cite{baque2017deep}            & 75.2    & 54.7        & 97.8     & 80.2  & 88.1   & 74.1    & 53.8        & 95     & 80  & 86.9
                                                  \\
    Volumetric \cite{iskakov2019learnable}   & 84.2    & 80.3        & 97.5     & 86.4  & 91.6   & 88.6    & 73.8        & 95.3     & 93.2  & 94.2
                                                  \\
\hline
    MVDet   \cite{hou2020multiview}          & 83.9    & 79.6        & 96.8        & 86.7  & 91.5     & 88.2    & 75.7        & 94.7        & 93.6  & 94.1
                                                  \\
    SHOT    \cite{song2021stacked}           & 88.3    & 82.0        & 96.6        & 91.5  & 94.0       & 90.2    & 76.5        & 96.1        & 94.0  & 95.0
                                                  \\
    MVDeTr  \cite{hou2021multiview}         & 93.7    & \textbf{91.3}        & \textbf{99.5}        & 94.2  & 97.8     & 91.5    & \textbf{82.1}        & \textbf{97.4}        & 94.0  & 95.7        \\
    3DROM   \cite{qiu20223d}     & 95.0    & 84.9        & 99.0       & 96.1  & 97.5    & \textbf{93.5}    & 75.9        & \underline{97.2}        & 96.2  & \textbf{96.7}                              \\

    EarlyBird \cite{Teepe_2024_WACV}     & 94.2    & \underline{90.1}     & 98.6      & 95.7   & 97.1    & 91.2 
                        & \underline{81.8}    & 94.9        & 96.3        & 95.6                        \\

    TrackTacular \cite{Teepe_2024_CVPR}     & \underline{96.5}    & 75.0        & \underline{99.4}       & 97.1  & \underline{98.2}    & \underline{92.1}    & 76.2       & 97.0        & 95.1  &   96.0                          \\
\hline
\hline
    E-MVOT (Ours)              & 96.3	& 85.2     & 98.1     & \textbf{98.1}      & {98.1}        & 91.9	& 80.9     & 94.1     & \textbf{98.0}      & 96.0    \\
    M-MVOT (Ours)               & \textbf{96.7}	& 86.1     & 98.8     & \underline{97.9}      & \textbf{98.3}           & \underline{92.1}	& 81.3    & 94.5     & \underline{97.8}     & \underline{96.1}           \\

\hline
\end{tabular}

\label{table:Wildtrack_MultiviewX_results}
\end{table*}

\textbf{Evaluation metrics.}
We use 5 metrics \cite{bernardin2008evaluating} to evaluate and compare the multi-view crowd localization methods:
Multiple Object Detection Accuracy (MODA), Multiple Object Detection Precision (MODP), Precision (P), Recall (R), and F1\_score (F1).
Specifically, 
$MODA = 1 - (FP+FN)/(TP+FN)$ measures the detection accuracy.
$MODP = (\sum(1-d[d<t]/t))/TP$ measures the precision of detection, where $d$ is the distance from a detected person point to its ground truth and $t$ is the distance threshold.
Meanwhile, 
$P = TP/(FP+TP)$, $R = TP/(TP+FN)$, and $F1 = 2P*R/(P+R)$.
Here, $TP$, $FP$, and $FN$ are the number of true positives, false positives, and false negatives, respectively.
Overall, \textbf{we use \textit{MODA} and \textit{F1\_score} as the main metrics for multi-view crowd localization}, since they comprehensively measure the performance of detecting the crowd both completely and precisely.

We use another 5 metrics \cite{bernardin2008evaluating} to evaluate and compare the performance of multi-view crowd tracking methods: Identification F1 score (IDF1), Multiple Object Tracking Accuracy (MOTA), Multiple Object Tracking Precision (MOTP), Mostly Tracked Targets (MT), and Mostly Lost Targets (ML).  
Specifically, \textit{IDF1} is the ratio of correctly identified detections over the average number of ground-truth and computed detections, which reflects the identity preservation capability.  
\textit{MOTA} measures the overall tracking accuracy by considering three types of errors: false positives, missed targets, and identity switches. It is defined as:  
\begin{equation}
    \text{MOTA} = 1 - \frac{FP + FN + IDS}{GT},
\end{equation}
where $FP$, $FN$, and $IDS$ are the numbers of false positives, false negatives, and identity switches, and $GT$ is the number of ground truth objects.  
\textit{MOTP} evaluates the average localization precision of correctly matched object trajectories across all frames and is defined as:
\begin{equation}
    \text{MOTP} = \frac{\sum_{i,t} d_{i}^{t}}{\sum_t c^t},
\end{equation}
where $d_i^t$ is the distance between the ground truth and predicted location of object $i$ at time $t$, and $c^t$ is the number of matches at time $t$.  
\textit{MT} is the percentage of ground-truth trajectories that are successfully tracked for more than 80\% of their lifespan, while \textit{ML} denotes the percentage of trajectories that are tracked for less than 20\% of their lifespan.  
Overall, \textbf{we use \textit{IDF1} and \textit{MOTA} as the main quantitative metrics for multi-view crowd tracking}, since they provide a comprehensive evaluation of both identity consistency and tracking accuracy.

\subsection{Multi-view crowd localization results}

\textbf{CVCS.}
In Table \ref{table:CVCS_results}, we compare the proposed Mahalanobis distance multi-view optimal transport loss (M-MVOT) with other multi-view crowd localization methods and the Euclidean distance multi-view optimal transport loss (E-MVOT) on the CVCS dataset. 
\red{For transparency, it reports both the ECCV 2024 results obtained with single-GPU training (batch size 1) and the journal-version results obtained with six-GPU distributed training (batch size 1 per GPU), with all other settings unchanged.}
Table \ref{table:CVCS_results} shows that the proposed M-MVOT loss achieves better overall localization performance than existing multi-view crowd localization methods. Even though MVDeTr achieves the highest MODP, our proposed M-MVOT is the best in terms of MODA and F1\_score, which are significantly increased from previous methods due to M-MVOT's higher recall rates. M-MVOT also outperforms E-MVOT, due to its ability to more flexibly change the cost function based on the relationships between the camera view and ground-truth positions.
Note that CVCS is a large multi-scene multi-view dataset, which is more challenging than other single-scene multi-view datasets.

\begin{figure*}[t]
\begin{center}
   \includegraphics[width=\linewidth]{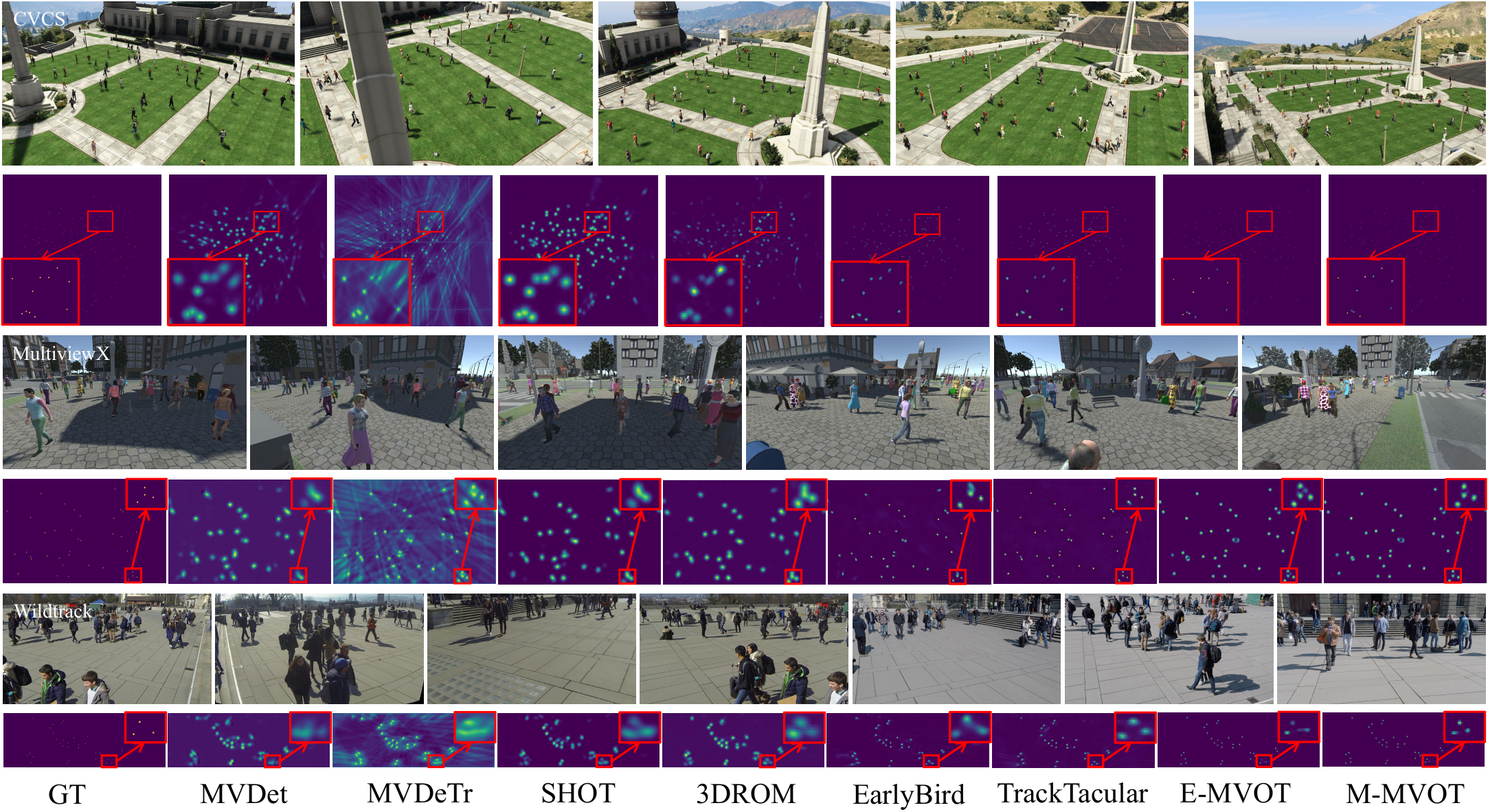}
\end{center}
\vspace{-0.3cm}
   \caption{The predicted crowd occupancy maps of different methods on the 3 datasets CVCS, MultiviewX, and Wildtrack (zoom in for better view). The value of crowd occupancy maps indicates the probability of a person existing at each location.}

\label{fig:results}
\vspace{-0.3cm}
\end{figure*}

\begin{table*}[t]
\centering
\caption{The ablation study on the MVOT loss for multi-view crowd localization on the CVCS dataset, with other settings kept the same except for the training loss.}
\begin{tabular}{l|cc|ccccc}
\hline
    Loss            & Dir.    & Dis.       & MODA$\uparrow$     & MODP$\uparrow$         & P$\uparrow$         & R$\uparrow$      & F1.$\uparrow$      \\
\hline
    MSE                &\ding{55}    &\ding{55}         &18.6     &74.4     &62.5      &46.5       &53.5    \\
    E-MVOT            & \ding{55}    & \ding{55}         &42.8 	&72.4      &76.7      &61.4     &68.2     \\
    MV-MVOT      & \ding{51}    & \ding{55}            &44.5   &73.1     &79.2      &60.3     & \textbf{68.5} \\
    ED-MVOT      & \ding{55}    & \ding{51}           &43.8     &\textbf{73.2}      &77.9      &\textbf{61.2}      & \textbf{68.5}  \\
    M-MVOT    & \ding{51}   & \ding{51}                   & \textbf{45.8}	  & \textbf{73.2}     & \textbf{82.0}    & 58.7      & 68.4   \\

\hline
\end{tabular}

\label{table:loss}
\end{table*}

\textbf{Wildtrack and MultiviewX.}
In Table \ref{table:Wildtrack_MultiviewX_results}, we compare our M-MVOT and E-MVOT with other multi-view crowd localization methods on the MultiviewX dataset (left) and Wildtrack dataset (right), respectively.
On \textbf{MultiviewX}, the proposed M-MVOT outperforms all comparison methods, achieving the highest MODA and F1 score. While E-MVOT performs slightly worse than TrackTacular, it still achieves competitive overall performance. Furthermore, M-MVOT surpasses E-MVOT, highlighting its advantage over the conventional Euclidean-distance-based optimal transport loss.
On \textbf{Wildtrack}, 3DROM achieves the best results on the MODA and F1\_score metrics, while M-MVOT obtains the second-best performance. A possible reason is that as a data augmentation method, 3DROM is more effective in alleviating overfitting on small-scale datasets such as Wildtrack. TrackTacular achieves similar performance to M-MVOT in terms of MODA, but performs worse than M-MVOT overall across all other metrics.
Nevertheless, M-MVOT still outperforms E-MVOT, demonstrating clearly the effectiveness of incorporating view ray direction and object-to-camera distance guidance in optimal transport framework.



\textbf{Visualization results.}
We show the corresponding visualization result on the 3 datasets in Fig. \ref{fig:results}.
Generally, the proposed M-MVOT and E-MVOT significantly show better localization performance under crowded regions  (in red boxes) and reduce the ground-plane projection artifacts of comparison methods in faraway regions, where density map-based methods tend to produce blurred or ambiguous responses.
Furthermore, as seen in the middle and the bottom of Fig. \ref{fig:results}, M-MVOT predicts fewer false positives than E-MVOT on MultiviewX dataset; M-MVOT also has notably fewer tail (streaking) artifacts than E-MVOT on Wildtrack dataset, leading to cleaner and more stable predictions, which proves the effectiveness of the proposed view ray direction and object-to-camera distance guidance in resolving localization ambiguities.

\begin{figure*}[t]
\begin{center}
   \includegraphics[width=\linewidth]{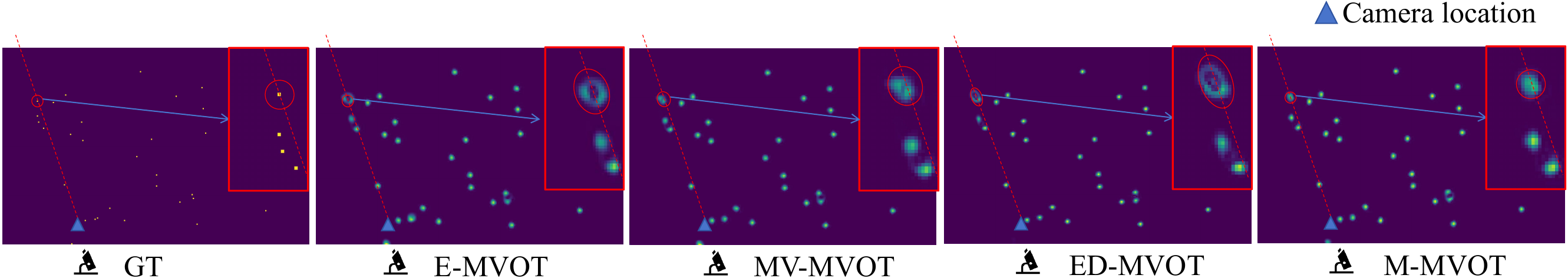}
\end{center}
\vspace{-0.2cm}
   \caption{Comparison of E-MVOT, MV-MVOT, ED-MVOT, and M-MVOT. The blue triangle is the camera location. M-MVOT predicts fewer artifacts than E-MVOT, MV-MVOT, and ED-MVOT (see red boxes), demonstrating the effectiveness of incorporating distance adjustment and view ray direction.}
\vspace{-0.3cm}
\label{fig:edmvot}
\end{figure*}

\begin{table*}[t]
\centering
\caption{The ablation study on $\alpha$ and $\tau$ of M-MVOT loss on the MultiviewX dataset. The best results are achieved by $\alpha=0.05$ and $\tau=20$.}

\begin{tabular}
{l|ccccc||l|ccccc}
\hline
    $\alpha$     & MODA$\uparrow$     & MODP$\uparrow$         & P.$\uparrow$         & R.$\uparrow$      & F1.$\uparrow$   
    & $\tau$     & MODA$\uparrow$     & MODP$\uparrow$         & P.$\uparrow$         & R.$\uparrow$      & F1.$\uparrow$   \\
\hline
   0      & 96.3    &  85.2    & 98.1     & \textbf{98.1}      & 98.1            
   &1      & 95.2    &  \textbf{90.2}    & 99.1     & 96.1      & 97.6  \\
   0.05      & \textbf{96.7}    & \textbf{ 86.1}    & \textbf{98.8}     & 97.9      & \textbf{98.3}    
   &10        & 96.1    & 86.5     & \textbf{99.2}     & 96.9      & 98.0   \\
   0.2        & \textbf{96.7}    & 84.3     & 98.6     & \textbf{98.1}      & \textbf{98.3}   &20     & \textbf{96.7}    & 86.1     & 98.8     & \textbf{97.9}      & \textbf{98.3}  \\
   0.6      & 96.3    & 84.5     & 98.2     & \textbf{98.1}      & 98.1    
   &30     & 95.2    & 83.8     & 97.9     & 97.3      & 97.6   \\
\hline
\end{tabular}
\label{table:alpha}
\end{table*}

\subsection{Ablation study on multi-view crowd localization}

\textbf{Ablation study on the MVOT loss.}
We conducted an ablation study on the setting of the MVOT loss in Table \ref{table:loss} on the CVCS dataset, comparing the proposed M-MVOT loss with MSE, MVOT with Euclidean cost (E-MVOT), MVOT with Mahalanobis cost based on view ray guidance only (MV-MVOT), and camera distance only (ED-MVOT). 
From Table \ref{table:loss}, M-MVOT achieves better performance than ED-MVOT, MV-MVOT, E-MVOT, and MSE loss, which demonstrates the advantage of the proposed distance guidance and view ray direction in the covariance matrix over other losses.
Note that ED-MVOT got lower results, due to more penalties being added to faraway points in all directions, leading to fewer predictions (so it has higher precision). M-MVOT considers directions and adds relaxations to directions perpendicular to the view ray, achieving better results. 
As shown in the red boxes of Fig. \ref{fig:edmvot}, M-MVOT predicts more accurate results than E-MVOT, MV-MVOT and ED-MVOT, which demonstrates the effectiveness of using the view ray direction and object-to-camera distance in the covariance matrix of the MVOT loss.



\textbf{Ablation study on $\alpha$ and $\tau$ in M-MVOT.}
We conducted an ablation study on the $\alpha$ and $\tau$ of M-MVOT in Table \ref{table:alpha} on MultiviewX.
$\alpha$ adjusts the object-to-camera distance in the M-MVOT loss in (\ref{eqn:md_dir_dis}). $\alpha=0$ is equivalent to the E-MVOT loss.
When $\alpha$ increases, the effect of the distance change on the model is enlarged, which penalizes the faraway regions for more accurate localization performance. However, when $\alpha$ is too large, the penalties are too large, which may force the model to neglect the near regions.
$\tau$ adjusts the effect of the last two regularization terms in (\ref{eqn:eot}) to ensure that all predicted density values have a corresponding annotation and all ground-truth points are used in the transport plan.
When $\tau$ increases, MODA and Recall increase while MODP and Precision drop for detecting more points. 
But when $\tau$ is too large, the penalties of the last two terms are too large, harming the optimization of the main term. 
The best result is achieved at $\alpha=0.05$, $\tau=20$.

\begin{table}[t]
\setlength{\tabcolsep}{3pt}
\small
\centering
\caption{The ablation study on $\sigma_1^2:\sigma_2^2$ of MV-MVOT loss on the MultiviewX dataset. The best results of fixed ratios are achieved by $\sigma_1^2:\sigma_2^2$ = 1:1.2, but it is still lower than M-MVOT's performance.}

\begin{tabular}{l|ccccc}
\hline
    $\sigma_1^2:\sigma_2^2$                 & MODA$\uparrow$     & MODP$\uparrow$         & P.$\uparrow$         & R.$\uparrow$      & F1.$\uparrow$      \\
\hline
   1:1      & 96.3    &  85.2    & 98.1     & 98.1      & 98.1    \\
\hline
   1.2:1     & 96.5    & 84.6   & 98.3     & \textbf{98.2}      & 98.2    \\
   1.5:1        & 96.2    & 85.2     & 98.2     & 98.0      & 98.1    \\
   2:1     & 96.3    & 86.1     & 98.5     & 97.8      & 98.1    \\

\hline
   1:1.2     & 96.5    & 85.6    & 98.3     & \textbf{98.2}     & 98.2    \\
   1:1.5        & 96.5    & 85.2     & 98.5     & 97.9      & 98.2    \\
   1:2     & 96.3    & 85.5     & 98.4     & 97.9      & 98.1    \\
\hline
   M-MVOT     & \textbf{96.7}    & \textbf{86.1}   & \textbf{98.8}     & 97.9      & \textbf{98.3}    \\
\hline
\end{tabular}
\label{table:sigma}
\vspace{-0.5cm}
\end{table}

\textbf{Ablation study on the fixed ratios of $\sigma_1^2$ and $\sigma_2^2$  }
We conducted more experiments on MV-MVOT in Table \ref{table:sigma} with various fixed ratios of $\sigma_1^2$ and $\sigma_2^2$ on the MultiviewX dataset. 
When $\sigma_1^2$ = $\sigma_2^2$ = 1, it is equal to E-MVOT.
When $\sigma_1^2>\sigma_2^2$ (1.2:1, 1.5:1, and 2:1), it means we encourage the prediction points to appear along the view ray direction, which is the opposite of MV-MVOT. 1.2:1 achieves the best performance with MODA=96.5, MODP=84.6, which is lower than the best performance of using $\sigma_1^2<\sigma_2^2$ (1:1.2), with MODA=96.5, MODP=85.6.
The reason is that the prediction points already appear along the view ray direction with a higher probability but with an offset to the ground-truth locations due to projection distortions. 
Thus, we enforce more punishments along the view ray direction to obtain more accurate predicted points, namely $\sigma_1^2$ should be smaller than $\sigma_2^2$ (1:1.2, 1:1.5, and 1:2).
From Table \ref{table:sigma}, we can see that the best performance on fixed ratio is achieved at $\sigma_1^2:\sigma_2^2$ = 1:1.2. When the ratio continues to increase, the performance starts to decrease. 

Overall, the best performance on fixed ratios is still worse than the proposed M-MVOT. Because M-MVOT considers the object-to-camera distance in the cost calculation and can adjust the cost adaptively and according to the distance to the camera.

\textbf{Ablation study on the multi-view loss fusion strategy.}
We conduct extra experiments on the multi-view fusion strategy with a very basic form on M-MVOT, that is, \textbf{Average}. We sum the OT losses under all cameras together and then average them. And \textbf{Soft} is instead to replace the binary choice strategy with soft weights learned from camera distance for fusing multiple cameras. The experiment in Table \ref{table:fusion} shows that the \textbf{distance-based selection} strategy achieves better results than Average or Soft fusion of multi-cameras, a simple but effective strategy for selecting the most reliable camera. In contrast, the average strategy considers all cameras equally without any preference, which is not sufficiently flexible for multi-view fusion.
\begin{table}[t]

\small
\centering
\caption{The ablation study on different strategies to fuse M-MVOT loss on the MultiviewX dataset. The best results are achieved by Dist-based Selection.}
\begin{tabular}{l@{\hskip 3pt}|c@{\hskip 2pt}c@{\hskip 2pt}c@{\hskip 2pt}c@{\hskip 2pt}c}
\hline
     Strategy                & MODA$\uparrow$    & MODP$\uparrow$        & P.$\uparrow$      & R.$\uparrow$     & F1.$\uparrow$       \\
\hline
   Average      & 96.5    &  85.2    & 98.7     & 97.8      & 98.2    \\
   Soft       & 96.5    & 86.0    & 98.7     &  97.8    & 98.2     \\
   Dist-based Sel.     & \textbf{96.7}    & \textbf{86.1}   & \textbf{98.8}   & \textbf{97.9}  & \textbf{98.3} \\
\hline
\end{tabular}
\label{table:fusion}
\vspace{-0.7cm}
\end{table}

\red{
\textbf{Comparison with learnable covariance matrix (LCM).}
We compare the proposed M-MVOT with a data-driven learnable covariance matrix (LCM) in Table \ref{table:lcm}. For LCM, the covariance matrices are estimated from the ground-plane features using two convolutional layers. As shown in Table \ref{table:lcm}, LCM does not achieve better performance than M-MVOT. The main reason is that M-MVOT explicitly incorporates the domain knowledge of camera geometry into the covariance matrix by considering the camera view direction and the object-to-camera distance. In contrast, LCM is optimized from the ground-plane features without explicit geometric constraints. Although LCM has the flexibility to adapt to the data distribution, it is difficult for the network to automatically discover and encode such view-dependent geometric priors during training. By explicitly modeling these geometry-aware factors, M-MVOT provides a more effective and interpretable covariance computation, making it more suitable for the multi-view localization task. Exploring how to incorporate geometric priors into learnable covariance matrices is an interesting direction for future work.}

\textbf{Comparison with the single-image SOTA OT method with a simple multi-view fusion (Multi-UOT).}
We compare our proposed M-MVOT with the single-image SOTA optimal transport method UOT \cite{wan2021generalized} with a simple multi-view fusion strategy.
Specifically, we project and sum the camera view results of \cite{wan2021generalized} on the ground plane and show results as Multi-UOT in Table \ref{table:lcm}.
 M-MVOT is much better than Multi-UOT, and it demonstrates that single-image methods with simple multi-view fusion strategy cannot fully utilize the multi-view information well, consistent with \cite{hou2020multiview, zhang2019wide}.

\begin{table}[t]
\setlength{\tabcolsep}{4pt}
\centering
\caption{Comparisons with learnable covariance matrix (LCM), and multi-view fusion of single-view methods (Multi-UOT).}
\begin{tabular}{l|ccccc}
\hline
    Comparison                  & MODA$\uparrow$     & MODP$\uparrow$         & P.$\uparrow$         & R.$\uparrow$      & F1.$\uparrow$      \\
\hline
    LCM                  & 95.1    & 81.5     & 96.7     &  \textbf{98.5}    & 97.6       \\
    Multi-UOT            & 23.9     & 72.0      & 75.2 & 35.7 & 48.4 \\
    M-MVOT               & \textbf{96.7}    & \textbf{86.1}     &  \textbf{98.8}    & 97.9      & \textbf{98.3}  \\

\hline
\end{tabular}
\vspace{-0.7cm}
\label{table:lcm}
\end{table}

\begin{table}[t]
\small
\centering
\setlength{\tabcolsep}{1pt}
\caption{The ablation study on the different backbones and losses on the MultiviewX dataset. The best results are achieved by MVDeTr with our M-MVOT loss.}
\begin{tabular}{l@{\hskip 0.2pt}|c@{\hskip 0.2pt}c@{\hskip 1pt}c@{\hskip 1pt}c@{\hskip 1pt}c@{\hskip 2pt}c}
\hline
Backbone    & Loss         & MODA$\uparrow$    & MODP$\uparrow$        & P.$\uparrow$      & R.$\uparrow$     & F1.$\uparrow$       \\
\hline
\multirow{3}{*}{MVDeTr}
    & MSE          & 93.9    & 90.3   & 98.0     & 95.9      & 96.9   \\
    & Focal        &94.5    &\textbf{93.3}   & \textbf{99.5}     &95.0     &97.2  \\
    & M-MVOT       & \textbf{96.7}    & 86.1   & 98.8     & \textbf{97.9}      & \textbf{98.3}      \\
\hline
\multirow{3}{*}{EarlyBird}
    & MSE         &92.6     &88.0      &97.7      &94.8       &96.2    \\
    & Focal         &94.2     &90.1      &98.6      &95.7       &97.1    \\
    & M-MVOT      &95.0     &87.7      &98.7    &96.3       &97.5      \\

\hline

\end{tabular}
\label{table:backbone}
\vspace{-0.7cm}
\end{table}

\textbf{Ablation study on different backbones and losses.}
We conducted experiments on our proposed M-MVOT with two different backbones on the MultiviewX dataset: MVDeTr \cite{hou2021multiview} and EarlyBird \cite{Teepe_2024_WACV}. In the MVDeTr backbone, we also add a multi-height selection module proposed by SHOT to reduce the impact of projection errors. 
The experiment in Table \ref{table:backbone} shows that the MVDeTr backbone performs better than the EarlyBird backbone.
More importantly, no matter which backbone it is implemented on, the proposed M-MVOT always achieves better performance than density-map-based MSE loss and Focal loss, which indicates the advantage of the proposed M-MVOT loss.



\begin{table}[t]
\small
\centering
\setlength{\tabcolsep}{1pt}
\caption{Ablation study on different dist. thresholds.}
\begin{tabular}{c|c|ccccc}
\hline

     Thres.      &Method          & MODA$\uparrow$    & MODP$\uparrow$        & P.$\uparrow$        & R.$\uparrow$     & F1.$\uparrow$     \\
\hline
\multirow{3}{*}{0.4m}
    &SHOT\cite{song2021stacked}  &89.0  &78.0   &97.1    &91.7        &94.3   \\
    &MVDeTr\cite{hou2021multiview}  &93.4  &\textbf{89.4}     &\textbf{99.4}    &93.9          &96.4   \\
    &M-MVOT &\textbf{96.5}    &82.7      &98.6     &\textbf{97.8}      &\textbf{98.2}     \\
\hline
\multirow{3}{*}{0.5m}
    &SHOT\cite{song2021stacked}  &88.3  &82.0    &96.6    &91.5        &94.0  \\
    &MVDeTr\cite{hou2021multiview}     &93.7      &\textbf{91.3}         &\textbf{99.5}        &94.2      &97.8   \\
    &M-MVOT      &\textbf{96.7}       &86.1        &98.8          &\textbf{97.9}       &\textbf{98.3}      \\
\hline
\multirow{3}{*}{0.6m}
    &SHOT\cite{song2021stacked}  &89.5  &85.1    &97.4    &92.0         &94.6  \\
    &MVDeTr\cite{hou2021multiview}    &93.6        &\textbf{92.9}         &\textbf{99.6}        &94.0      &96.7   \\
    &M-MVOT &\textbf{96.7}       &88.4          &98.8          &\textbf{97.9}       &\textbf{98.3}      \\

\hline
\end{tabular}
\vspace{-0.7cm}
\label{table:distthresld}
\end{table}

\textbf{Ablation study on different distance thresholds.}
In addition to using a 0.5 m distance threshold, we further evaluate and compare results under different thresholds in Table \ref{table:distthresld}. The 0.5 m setting is reasonable given human body size, and is also adopted by all SOTA methods \cite{hou2020multiview,hou2021multiview,song2021stacked,qiu20223d}. We additionally evaluate thresholds of {0.4 m, 0.6 m} on MultiviewX, as shown in Table \ref{table:distthresld}. Our method consistently outperforms MVDeTr \cite{hou2021multiview} and SHOT \cite{song2021stacked} in terms of MODA and F1 score, demonstrating that the proposed M-MVOT loss remains stable across different thresholds and is robust to threshold selection.

\begin{table}[t]
\small
\centering
\setlength{\tabcolsep}{2.5pt}
\caption{Computation cost comparison among different multi-view localization loss functions.}
\vspace{-0.2cm}
\begin{tabular}{l|cccc}
\hline
Loss & Mem.(GB) & FLOPs(G) & Train(s) & Test(s) \\
\hline
MSE    &7.446  &1224.243  &0.595  &0.121  \\
Focal    & 7.446 & 1224.243 & 0.583 & 0.121 \\
E-MVOT    & 7.446 & 1224.243 & 0.612 & 0.121 \\
M-MVOT  & 7.448 & 1224.243 & 0.642 & 0.121 \\
\hline
\end{tabular}
\label{table:computation_loss}
\vspace{-0.9cm}
\end{table}

\begin{table}[t]
\small
\centering
\setlength{\tabcolsep}{2pt}
\caption{Computation cost comparison among different multi-view localization methods.}
\vspace{-0.2cm}
\begin{tabular}{l|cccc}
\hline
Method & Mem.(GB) & FLOPs(G) & Tr.(s) & Te.(s) \\
\hline
MVDet\cite{hou2020multiview}      & 12.633 & 2032.899 & 0.309 & 0.274 \\
MVDeTr\cite{hou2021multiview}     & 19.430 & 1269.489 & 0.259 & 0.225 \\
SHOT\cite{song2021stacked}        & 6.070 & 2040.674 & 0.412 & 0.308 \\
3DROM\cite{qiu20223d}             & 14.387 & 4017.931 & 0.609 & 0.235 \\
EarlyBird\cite{Teepe_2024_WACV}   & 5.880 & 1074.787 & 0.314 & 0.064 \\
TrackTacular\cite{Teepe_2024_CVPR}& 9.838 & 1221.315 & 0.298 & 0.068 \\
M-MVOT                     & 7.448 & 1224.243 & 0.642 & 0.121 \\
\hline
\end{tabular}
\label{table:computation_methods}
\vspace{-0.7cm}
\end{table}

\textbf{Ablation study on the computational cost of different losses on the multi-view localization task.}
\red{We compare the computation costs for different losses in Table \ref{table:computation_loss}, where all methods share the same network architecture. Therefore, the reported FLOPs are identical because they measure only the network forward pass and exclude loss computation. GPU memory consumption is also nearly identical, as it is dominated by the intermediate feature maps and activations required for back-propagation, while the additional tensors introduced by different losses are negligible. Specifically, E-MVOT introduces almost no additional memory overhead, while M-MVOT incurs only a slight increase (from 7.446 GB to 7.448 GB) due to the extra Mahalanobis distance computation. The computational overhead of E-MVOT and M-MVOT is mainly reflected in training time, while test-time speed remains unchanged since the loss functions are not used during inference.}

\textbf{Ablation study on the computational cost of different methods on the multi-view localization task.}
In Table \ref{table:computation_methods}, we compare the computation costs of our M-MVOT with MVDet \cite{hou2020multiview}, MVDeTr \cite{hou2021multiview}, SHOT \cite{song2021stacked}, 3DROM \cite{qiu20223d}, EarlyBird \cite{Teepe_2024_WACV}, and TrackTacular \cite{Teepe_2024_CVPR}. Among them, MVDeTr \cite{hou2021multiview} has the highest GPU memory consumption, while 3DROM \cite{qiu20223d} requires the largest number of FLOPs. In contrast, M-MVOT achieves the lowest FLOPs except TrackTacular and EarlyBird, indicating superior computational efficiency. EarlyBird and TrackTacular exhibit shorter training and testing times, likely due to their implementation with PyTorch Lightning, which provides more optimized training routines than standard PyTorch.

\begin{table}[!t]
\centering
\setlength{\tabcolsep}{1pt}
\caption{Comparison of the multi-view crowd tracking performance on MultiviewX. MVOT-Track achieves the best performance.}
\vspace{-0.2cm}

\begin{tabular}{l|ccccc}
\hline
Tracking Method & IDF1$\uparrow$ & MOTA$\uparrow$ & MOTP$\uparrow$ & MT$\uparrow$ & ML$\downarrow$ \\
\hline
\red{MVDet-TBD} \cite{hou2020multiview} &65.2 &69.0 &81.2 &44.7 &7.9 \\
\red{SHOT-TBD} \cite{song2021stacked} &74.9 &80.7 &83.1 &67.1 &6.6 \\
\red{MVDeTr-TBD} \cite{hou2021multiview} &74.4 &85.3 &85.5 &78.9 &3.9 \\
\red{3DROM-TBD} \cite{qiu20223d}  &74.5 &86.4 &83.1 &76.3 &3.9 \\
\hline
MVOT-TBD (Ours) &81.3 &90.9 &84.3 &89.5 & \underline{1.3} \\
\hline
REMP \cite{11018739} &– &81.0 &85.8 &– &– \\
MVTr \cite{yang2024end} &82.9 &91.4 &\textbf{95.0} &\underline{96.1} & \textbf{0.0} \\
EarlyBird \cite{Teepe_2024_WACV} & 82.4 & 88.4 & \underline{86.2} & 82.9 & \underline{1.3} \\
TrackTacular \cite{Teepe_2024_CVPR} & 85.6 & 92.4 & 80.1 & 92.1 & 2.6 \\
MVTrajecter \cite{Yamane_2025_ICCV}  &\underline{85.8} &\underline{92.8} &\textbf{95.0} &\textbf{97.4} &\textbf{0.0}\\
\hline
MVOT-Track (Ours) & \textbf{86.4} & \textbf{93.1} & 85.1 & 93.4 & \underline{1.3} \\
\hline
\end{tabular}
\label{table:tracking_multiviewx}
\vspace{-0.7cm}
\end{table}

\begin{table}[t]
\centering
\setlength{\tabcolsep}{1pt}
\caption{Comparison of multi-view crowd tracking performance on the Wildtrack dataset. 
$^{\dagger}$ results recomputed by EarlyBird \cite{Teepe_2024_WACV}. 
$^{*}$ denotes results reproduced by us for MVTrajecter \cite{Yamane_2025_ICCV}, 
as no official implementation is publicly available.}
\begin{tabular}{l|ccccc}
\hline
Tracking Method & IDF1$\uparrow$ & MOTA$\uparrow$ & MOTP$\uparrow$ & MT$\uparrow$ & ML$\downarrow$ \\
\hline
\red{MVDet-TBD} \cite{hou2020multiview} &85.5 &81.5 &83.3 &53.7 &17.1 \\
\red{SHOT-TBD} \cite{song2021stacked} &89.3 &86.0 &85.6 &68.3 &12.2\\ 
\red{MVDeTr-TBD} \cite{hou2021multiview} &91.1 &89.9 &88.3 &75.6 &\textbf{4.9}\\ 
\red{3DROM-TBD} \cite{qiu20223d} &91.4 &89.0 &83.8 &70.7 &\underline{7.3}\\
\hline
MVOT-TBD (Ours) &92.4 &90.4 &87.7 &87.8 &\textbf{4.9}\\
\hline
KSP-DO \cite{chavdarova2018wildtrack} & 73.2 & 69.6 & 61.5 & 28.7 & 25.1 \\
KSP-DO-ptrack \cite{chavdarova2018wildtrack} & 78.4 & 72.2 & 60.3 & 42.1 & 14.6 \\
GLMB-YOLOv3 \cite{ong2020bayesian} & 74.3 & 69.7 & 73.2 & 79.5 & 21.6 \\
GLMB-DO \cite{ong2020bayesian} & 72.5 & 70.1 & 63.1 & 93.6 & 22.8 \\
DMCT \cite{you2020real} & 77.8 & 72.8 & 79.1 & 61.0 & \textbf{4.9} \\
DMCT Stack \cite{you2020real} & 81.9 & 74.6 & 78.9 & 65.9 & \textbf{4.9} \\
ReST$^{\dagger}$ \cite{Cheng_2023_ICCV} & 86.7 & 84.9 & 84.1 & 87.8 & \textbf{4.9} \\
MVFlow \cite{engilberge2023multi} & 93.5 & 91.3 & - & - & - \\
MVTr \cite{yang2024end} &93.1 &\underline{92.3} &\underline{92.7} &\textbf{95.1} &\textbf{4.9} \\
EarlyBird \cite{Teepe_2024_WACV} & 92.3 & 89.5 & 86.6 & 78.0 & \textbf{4.9} \\
TrackTacular \cite{Teepe_2024_CVPR} & 95.3 & 91.8 & 85.4 & 87.8 & \textbf{4.9} \\
MVTrajecter \cite{Yamane_2025_ICCV}  &\textbf{96.5} &\textbf{94.3} &\textbf{93.0} &90.2 &\textbf{4.9}  \\
MVTrajecter* \cite{Yamane_2025_ICCV}  &89.3 &82.2 &87.8 &\underline{92.6} &\textbf{4.9}\\
\hline
MVOT-Track (Ours) & \underline{95.4} & 91.1 & 87.6 & 87.8 & \textbf{4.9} \\
\hline
\end{tabular}
\label{table:tracking_wildtrack}
\vspace{-0.7cm}
\end{table}

\begin{figure*}[t]
\begin{center}
   \includegraphics[width=\linewidth]{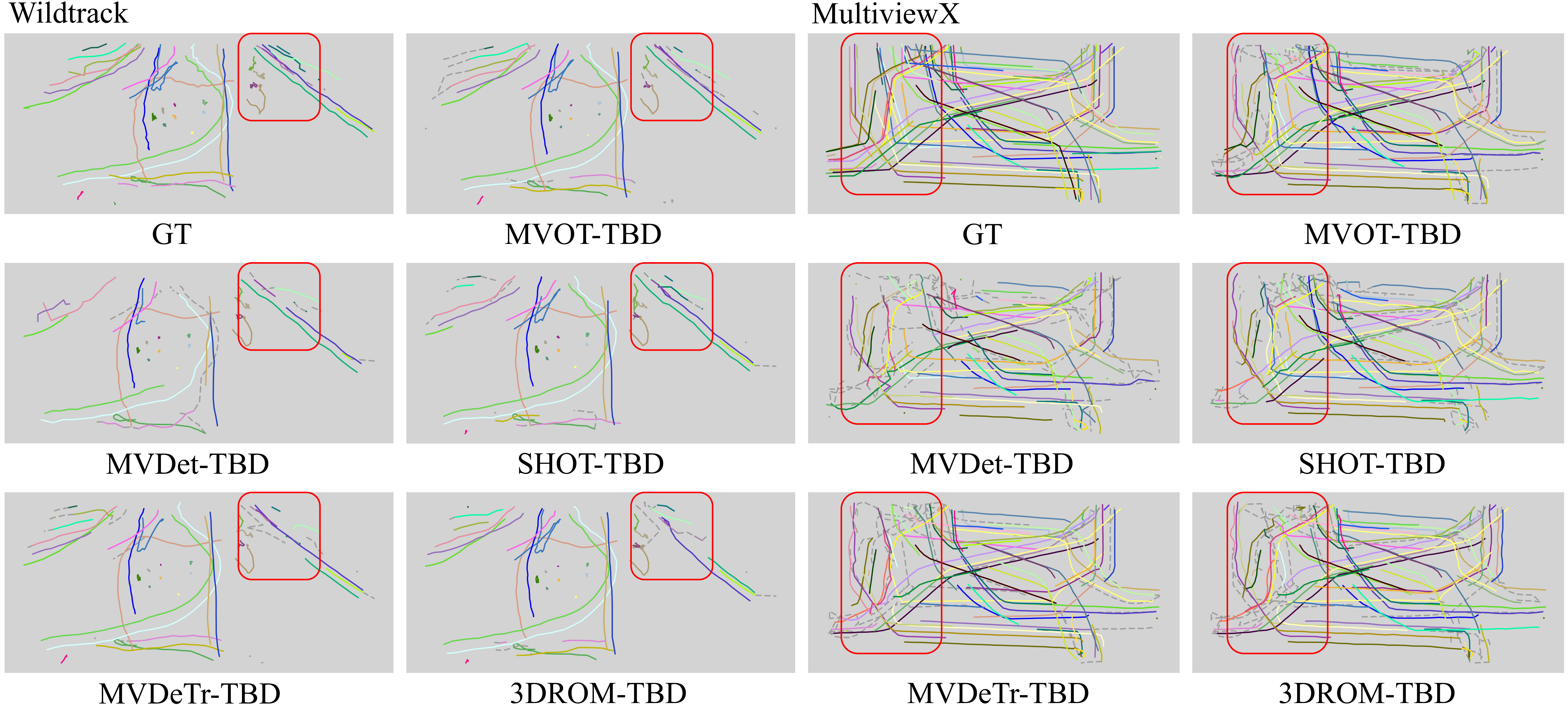}
\end{center}
\caption{\red{Qualitative comparison of MVOT-TBD with the tracking variants of multi-view localization methods on Wildtrack and MultiviewX (zoom in for better view). Colored trajectories indicate IDTPs, gray dashed trajectories indicate IDFPs, and red boxes highlight regions where MVOT-TBD produces fewer unmatched and fragmented trajectories.}}
\label{fig:visualization_tbd}
\end{figure*}

\subsection{Multi-view crowd tracking results}

In Tables \ref{table:tracking_multiviewx} and \ref{table:tracking_wildtrack}, we compare our proposed MVOT-TBD and MVOT-Track with other SOTA multi-view crowd tracking methods, as well as \red{the tracking variants constructed from the multi-view crowd localization methods} MVDet \cite{hou2020multiview}, SHOT \cite{song2021stacked}, MVDeTr \cite{hou2021multiview}, and 3DROM \cite{qiu20223d}. To ensure a fair and meaningful comparison, all compared localization methods are equipped with the same tracking pipeline as MVOT-TBD, \red{and the resulting variants are denoted as MVDet-TBD, SHOT-TBD, MVDeTr-TBD, and 3DROM-TBD, respectively. The suffix ``-TBD'' indicates that the corresponding localization method is integrated into the same tracking-by-detection pipeline.} As shown, MVOT-TBD achieves better performance than \red{MVDet-TBD, SHOT-TBD, MVDeTr-TBD, and 3DROM-TBD} in terms of IDF1, MOTA, MT, and ML on both the MultiviewX and Wildtrack datasets. These results \red{demonstrate the effectiveness of MVOT-TBD as a complete tracking-by-detection framework and also provide complementary evidence for the advantage of the proposed M-MVOT localization method under the same tracking pipeline.}

Furthermore, MVOT-Track achieves the highest IDF1 and MOTA on the MultiviewX dataset among all compared methods, outperforming MVTrajecter. This may be attributed to the fact that, although MVTrajecter leverages both motion and appearance cues, it assigns a very low weight to motion (0.02) and thus relies heavily on appearance (0.98). As MultiviewX is a synthetic dataset with limited appearance variations, such as illumination changes and sensor noise, it is more challenging to learn identity-consistent appearance embeddings than in real-world scenarios.
On Wildtrack, MVTrajecter reports the best performance in terms of IDF1 and MOTA, while MVOT-Track achieves competitive results. A possible explanation for this gap is that MVTrajecter incorporates longer temporal appearance context (up to 7 frames), which may better exploit temporal consistency. However, since the official implementation is not publicly available, we provide our own reproduction, which yields noticeably lower performance. This discrepancy suggests that the reported results may be sensitive to implementation details.
Overall, the performance gains of MVOT-Track on both datasets can be attributed to the explicit supervision of cross-frame associations, which results in fewer identity switches, better preservation of long-term trajectories, and reduced track fragmentation. 

\begin{figure*}[t]
\begin{center}
   \includegraphics[width=\linewidth]{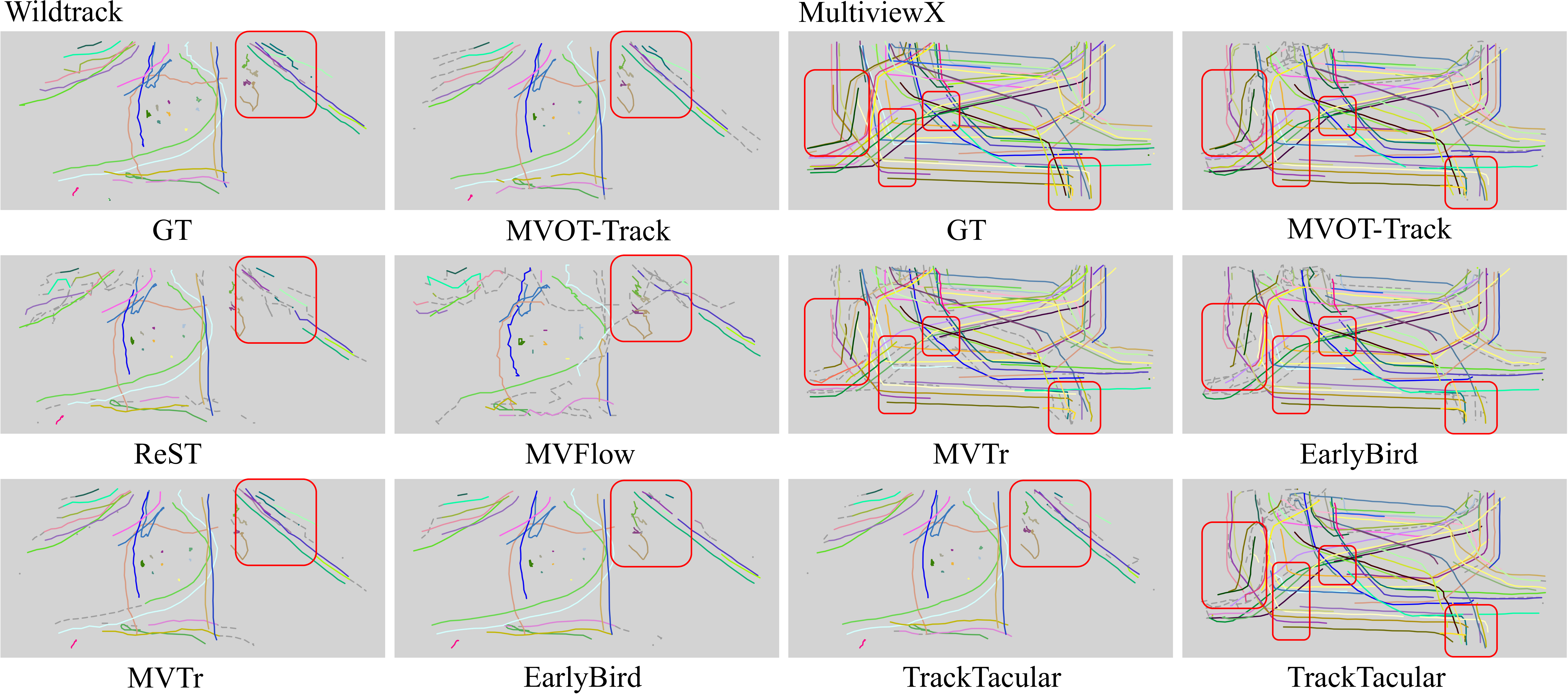}
\end{center}
\caption{\red{Qualitative comparison of MVOT-Track with state-of-the-art multi-view tracking methods on Wildtrack and MultiviewX (zoom in for better view). Colored trajectories indicate IDTPs, gray dashed trajectories indicate IDFPs, and red boxes highlight challenging regions where MVOT-Track recovers more complete trajectories with fewer tracking errors.}}
\label{fig:visualization_main}
\end{figure*}

\red{As shown in Figures~\ref{fig:visualization_tbd} and~\ref{fig:visualization_main}, we provide qualitative comparisons on both Wildtrack and MultiviewX. Colored trajectories denote correctly matched trajectories (IDTPs), while gray dashed trajectories denote unmatched predicted trajectories (IDFPs). Figure~\ref{fig:visualization_tbd} compares MVOT-TBD with the tracking variants of representative multi-view localization methods, including MVDet-TBD, SHOT-TBD, MVDeTr-TBD, and 3DROM-TBD. Figure~\ref{fig:visualization_main} further compares MVOT-Track with dedicated multi-view tracking methods. 
Overall, MVOT-TBD and MVOT-Track recover more trajectories that closely align with the ground truth while producing fewer gray dashed false-positive trajectories, particularly in the regions highlighted by the red boxes. Compared with MVOT-TBD, MVOT-Track further exhibits more complete and temporally coherent trajectories with fewer visible association failures, which is consistent with its higher IDF1 and MOTA on both datasets. The advantage is most evident in areas with closely spaced parallel motion, dense trajectory intersections, and scene entrances or exits, where competing methods more frequently generate fragmented or unmatched trajectories or terminate valid tracks prematurely. These qualitative results are consistent with the quantitative results and further demonstrate the effectiveness of the proposed methods in crowded regions.}

\begin{figure*}[t]
\begin{center}
   \includegraphics[width=\linewidth]{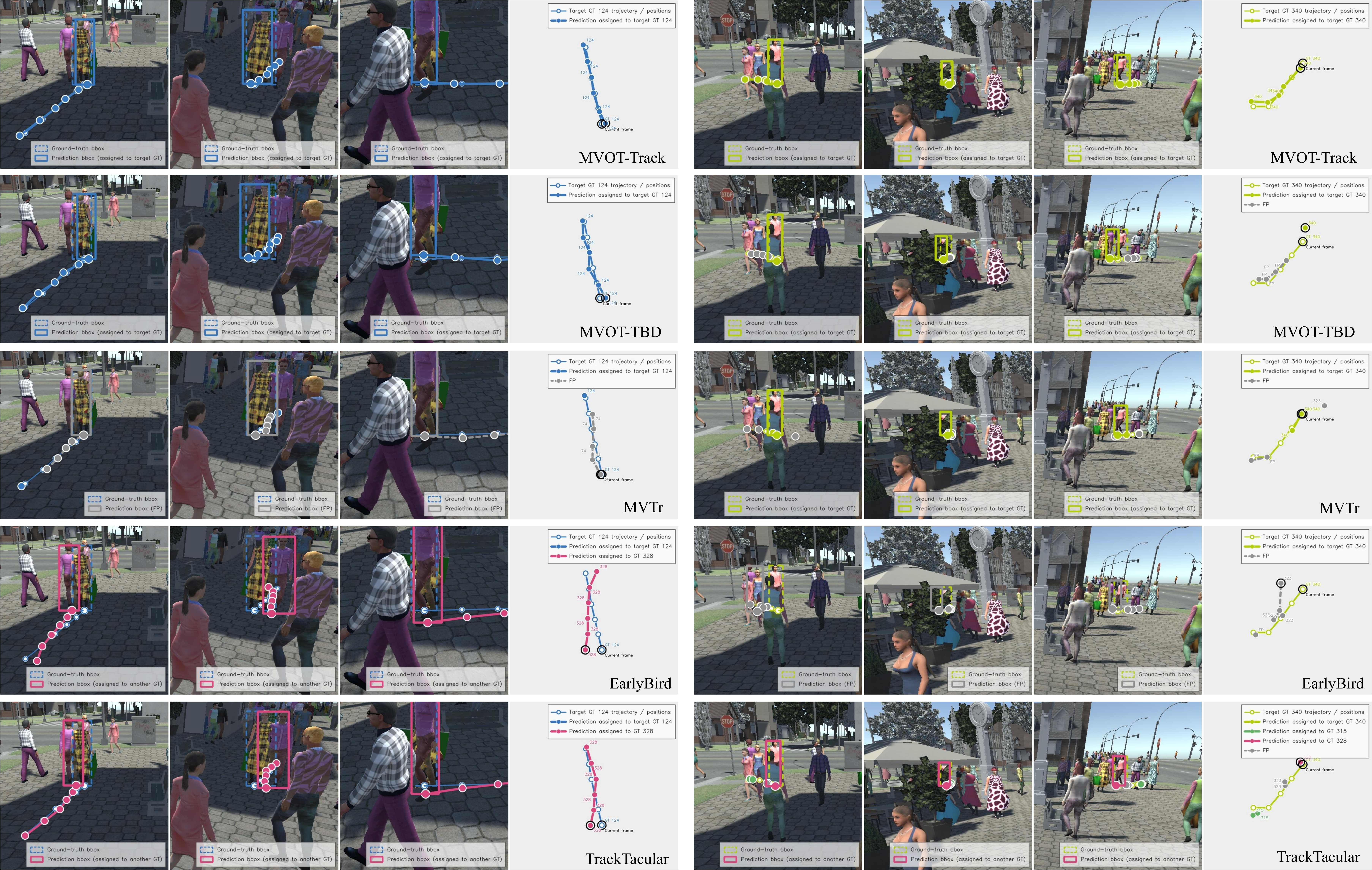}
\end{center}
\caption{\red{Cross-view tracking visualizations on MultiviewX (zoom in for better view). Each example shows the synchronized camera views in which the target pedestrian is visible, together with the corresponding BEV trajectories. Predictions associated with the target GT use the same color, associations to other GTs use their corresponding colors, and false positives are shown in gray. The numbers beside the BEV nodes indicate the associated GT IDs, and the current frame is marked in the BEV view.}}
\label{fig:visualization_cross_view}
\end{figure*}

\red{We further provide cross-view visualization examples on MultiviewX in Figure~\ref{fig:visualization_cross_view}. For each example, we select a target GT trajectory and the synchronized views where it is visible, and project the corresponding predicted ground-plane trajectories back using the camera parameters. The GT trajectory is shown with hollow markers and a dashed current-frame bounding box, while predictions use filled markers and solid bounding boxes. The last column presents the corresponding BEV trajectories, with numbers indicating the associated GT IDs. MVOT-Track maintains consistent identity associations and closely follows the target trajectory across views in both examples. Compared with MVOT-TBD, it follows the target more closely and avoids additional false-positive trajectory portions, while the other methods exhibit incorrect or multiple identity associations or prematurely lose the target. These results provide direct qualitative evidence of the improved temporal consistency of MVOT-Track in crowded scenes.}


\subsection{Ablation study on multi-view crowd tracking}

\textbf{Ablation study on $\gamma$ and $\psi$ in MVOT-Track.}
We perform an ablation study on the hyperparameters $\gamma$ and $\psi$ of MVOT-Track on the Wildtrack dataset, as shown in Table~\ref{table:gamma_psi}.  
$\gamma$ controls the weight of the marginal constraints on the transport matrix $\mathbf{P}^{tra}$ in Eq.~(\ref{mvot_track_formula}); setting $\gamma=0$ effectively removes these constraints.  
$\psi$ governs the strength of association correspondence supervision, i.e., the direct supervision imposed on $\mathbf{P}^{tra}$ to guide the alignment of tracks across frames. 
From Table~\ref{table:gamma_psi}, we observe that strong marginal constraints ($\gamma=10$) combined with moderate association supervision ($\psi=1$) achieve the best tracking performance, balancing IDF1, MOTA, and MT metrics.



\begin{table}[t]
\centering
\setlength{\tabcolsep}{7pt}
\caption{Ablation study on $\gamma$ and $\psi$ of MVOT-Track on the Wildtrack dataset. The best results are achieved when $\gamma=10$ and $\psi=1$.}
\label{table:gamma_psi}

\begin{tabular}{l|ccccc}
\hline
$\gamma$ & IDF1$\uparrow$ & MOTA$\uparrow$ & MOTP$\uparrow$ & MT$\uparrow$ & ML$\downarrow$ \\
\hline
0  & 95.1 & 90.1 & 87.1 & 85.4 & \textbf{4.9} \\
5  & 95.0 & 90.2 & \textbf{87.8} & \textbf{87.8} & \textbf{4.9} \\
10 & \textbf{95.4} & \textbf{91.1} & 87.6 & \textbf{87.8} & \textbf{4.9} \\
20 & 95.0 & 90.5 & 87.3 & 85.4 & \textbf{4.9} \\
\hline
\end{tabular}

\vspace{0.2cm}

\begin{tabular}{l|ccccc}
\hline
$\psi$ & IDF1$\uparrow$ & MOTA$\uparrow$ & MOTP$\uparrow$ & MT$\uparrow$ & ML$\downarrow$ \\
\hline
0   & \textbf{95.4} & 90.0 & 86.9 & 82.9 & \textbf{4.9} \\
0.1 & 95.0 & 89.0 & 86.9 & 82.9 & \textbf{4.9} \\
1   & \textbf{95.4} & \textbf{91.1} & \textbf{87.6} & \textbf{87.8} & \textbf{4.9} \\
3   & 95.2 & 89.2 & 87.1 & 80.5 & 7.3 \\
\hline
\end{tabular}
\vspace{-0.7cm}
\end{table}




\begin{table}[t]
\small
\centering
\setlength{\tabcolsep}{2pt}
\caption{\red{Ablation study on different tracking loss settings on the Wildtrack and MultiviewX datasets. Incorporating the proposed MVOT-Track loss yields clear improvements on both datasets.}}
\begin{tabular}{l|ccccc}
\hline
Loss & IDF1$\uparrow$ & MOTA$\uparrow$ & MOTP$\uparrow$ & MT$\uparrow$ & ML$\downarrow$ \\
\hline
\multicolumn{6}{c}{Wildtrack} \\
\hline
$L_1$
& 95.1
& 89.7
& 86.9
& \textbf{87.8}
& \textbf{4.9} \\[4pt]

\makecell[l]{MVOT-Track \\ + $L_1$ (Eq.~\ref{eqn:l_tracking})}
& \textbf{95.4}
& \textbf{91.1}
& \textbf{87.6}
& \textbf{87.8}
& \textbf{4.9} \\
\hline
\multicolumn{6}{c}{MultiviewX} \\
\hline
$L_1$
& 85.2
& 92.7
& \textbf{85.3}
& 89.5
& 2.6 \\[4pt]

\makecell[l]{MVOT-Track \\ + $L_1$ (Eq.~\ref{eqn:l_tracking})}
& \textbf{86.4}
& \textbf{93.1}
& 85.1
& \textbf{93.4}
& \textbf{1.3} \\
\hline
\end{tabular}
\vspace{-0.7cm}
\label{table:ablation_MVOT_Track_l1}
\end{table}

\red{\textbf{Ablation study on different tracking loss settings.}
Table \ref{table:ablation_MVOT_Track_l1} reports the ablation results of different tracking loss settings on Wildtrack and MultiviewX. We compare the proposed MVOT-Track combined with the standard $L_1$ loss against using the $L_1$ loss alone. On Wildtrack, introducing MVOT-Track improves IDF1, MOTA, and MOTP by 0.3, 1.4, and 0.7 points, respectively. On MultiviewX, it improves IDF1 and MOTA by 1.2 and 0.4 points, respectively, while increasing MT by 3.9 points and reducing ML by 1.3 points. MOTP remains comparable, with a slight decrease. Unlike the $L_1$ loss, which only supervises motion-offset regression, MVOT-Track explicitly supervises the temporal transport matrix and association correspondences between consecutive frames. These results demonstrate that MVOT-Track provides effective complementary supervision to the standard $L_1$ loss on both datasets.}

\begin{table}[t]
\small
\centering
\setlength{\tabcolsep}{1pt}
\caption{\red{Ablation study on the number of temporal frames on Wildtrack. Two frames achieve the best trade-off between tracking performance and computational cost.}}
\label{table:ablation_numHist}
\begin{tabular}{c|ccccc}
\hline
\multicolumn{6}{c}{Tracking performance} \\
\hline
\#Frames
& IDF1$\uparrow$
& MOTA$\uparrow$
& MOTP$\uparrow$
& MT$\uparrow$
& ML$\downarrow$ \\
\hline
1 & 94.4 & 89.5 & \textbf{87.9} & 82.9 & \textbf{4.9} \\
2 & \textbf{95.4} & \textbf{91.1} & 87.6 & \textbf{87.8} & \textbf{4.9} \\
3 & 94.6 & 89.6 & 87.4 & 85.3 & \textbf{4.9} \\
4 & 94.7 & 89.3 & \textbf{87.9} & 85.3 & \textbf{4.9} \\
\hline
\multicolumn{6}{c}{\red{Computational cost}} \\
\hline
\#Frames
& Mem. (GB)
& FLOPs (G)
& Tr. (s)
& \multicolumn{2}{c}{Te. (s)} \\
\hline
1 & 20.581 & 1468.210 & 0.671 & \multicolumn{2}{c}{0.494} \\
2 & 24.630 & 3062.921 & 1.184 & \multicolumn{2}{c}{0.580} \\
3 & 32.872 & 4568.450 & 1.666 & \multicolumn{2}{c}{0.728} \\
4 & 40.537 & 6073.980 & 2.206 & \multicolumn{2}{c}{0.885} \\
\hline
\end{tabular}
\vspace{-0.7cm}
\end{table}

\red{\textbf{Ablation study on the number of frames.}
We conduct an ablation study to investigate the effect of the number of temporal frames on both tracking performance and computational cost. In the default setting, we use two frames, namely the current frame $t$ and the previous frame $t-1$. As shown in Table~\ref{table:ablation_numHist}, this setting achieves the best IDF1, MOTA, and MT while maintaining moderate memory consumption, FLOPs, and runtime. Using only the current frame requires less computation but leads to lower tracking performance, demonstrating the benefit of incorporating temporal information. Increasing the number of frames to three or four substantially increases the computational cost without further improving tracking performance. These results indicate that two-frame temporal fusion provides the best trade-off between tracking effectiveness and computational efficiency.
}

\red{
\textbf{Ablation study on the number of camera views.}
We evaluate the robustness of MVOT-Track under different numbers of camera views on Wildtrack using nested subsets obtained by removing one camera at each step, as shown in Table~\ref{tab:view_ablation}. MVOT-Track remains robust under moderate view reduction: the six-view setting achieves the same IDF1 and comparable MOTA to the full seven-view setting, while the five-view setting still obtains an IDF1 of 94.5 and a MOTA of 89.4. More noticeable degradation occurs with four or three views because of reduced scene coverage and complementary cross-view information. The different trends of the three- and four-view subsets further indicate that performance depends on both the number and configuration of cameras.
}

\begin{table}[t]
\centering
\setlength{\tabcolsep}{1pt}
\caption{Ablation study on the number of camera views on the Wildtrack dataset. The model is retrained using different camera subsets.}
\label{tab:view_ablation}
\begin{tabular}{c|c|ccccc}
\hline
\# Views & Camera IDs  & IDF1$\uparrow$ & MOTA$\uparrow$ & MOTP$\uparrow$ & MT$\uparrow$ & ML$\downarrow$ \\
\hline

7 & 1,2,3,4,5,6,7 &\textbf{95.4} & 91.1 & \textbf{87.6} & \textbf{87.8} & \textbf{4.9} \\
6 & 1,2,3,5,6,7   &\textbf{95.4} &\textbf{91.4} &86.6 &82.9 &\textbf{4.9} \\
5 & 2,3,5,6,7     &94.5 &89.4 &86.4 &82.9 &7.3 \\
4 & 2,5,6,7       &91.6 &87.8 &84.9 &78.0 &7.3 \\
3 & 2,6,7         &93.3 &85.2 &84.6 &78.0 &9.7 \\

\hline

\end{tabular}
\vspace{-0.7cm}
\end{table}

\textbf{Ablation study on the computational cost of different methods on multi-view tracking.}
Table \ref{table:computation_tracking} compares the computational cost of different multi-view tracking methods. \red{ReST achieves the lowest memory consumption, FLOPs, and training time because it operates on detection-level features with lightweight spatial and temporal graph modules, while the cost of its external detector is excluded.}
MVFlow achieves the fastest inference speed, which is likely attributed to its lower input and ground-plane resolutions. 
While MVOT-TBD and MVOT-Track are slightly slower, they remain more efficient than TrackTacular and achieve the best overall tracking performance among all methods.


\begin{table}[t]
\small
\centering
\setlength{\tabcolsep}{1pt}
\caption{Computation cost comparison among different multi-view tracking methods. }

\begin{tabular}{l|cccc}
\hline
Method & Mem.(GB) & FLOPs(G) & Tr.(s) & Te.(s) \\
\hline
\red{ReST} \cite{Cheng_2023_ICCV} &\red{1.908}  &\red{134.958}  &\red{0.064}  &\red{0.295}  \\
\red{MVTr} \cite{yang2024end} &\red{16.101}  &\red{1269.498}  &\red{0.751}  &\red{0.142}  \\
MVFlow \cite{engilberge2023multi} & 8.717 & 1055.347 & 0.424 & 0.075 \\
EarlyBird\cite{Teepe_2024_WACV}   &6.737  &1220.915  &0.477  &0.150  \\
TrackTacular\cite{Teepe_2024_CVPR}&10.828  &2031.936  &0.728  & 0.609 \\
MVOT-TBD                     &22.890 &3062.920  &0.775  &0.465  \\
MVOT-Track  &24.630  &3062.921  &1.184  &0.580  \\
\hline
\end{tabular}
\vspace{-0.7cm}
\label{table:computation_tracking}
\end{table}

\section{Discussion and Conclusion}
\label{sec:conclusion}
\red{In this paper, we introduce M-MVOT, a novel optimal transport loss for point-supervised multi-view crowd localization. It incorporates camera view directions and object-to-camera distances into the Mahalanobis transport cost and applies the cost from the closest camera to reduce ground-plane projection distortions. We further extend the framework to tracking through MVOT-TBD, a complete tracking-by-detection pipeline, and MVOT-Track, a tracking-specific OT formulation that directly supervises the temporal transport plan and its marginals across consecutive frames. Extensive experiments demonstrate the effectiveness of M-MVOT for localization and the competitive tracking performance of MVOT-TBD and MVOT-Track.}

\section*{Acknowledgments}
This work was supported in part by National Key Research and Development Program of China (2024YFB3908500, 2024YFB3908504), National Natural Science Foundation of China (W2611096), Shenzhen Science and Technology Program (KQTD20210811090044003, KJZD20240903100022028), Scientific Foundation for Youth Scholars from Shenzhen University, City University of Hong Kong (Project No. 7030010), and Guangdong Provincial Key Laboratory of Visual Media and Multidimensional Intelligence (2026B1212010012).



\section*{Declarations}

\textbf{Ethical consideration}. Multi-view crowd localization and tracking could be used for crowd analysis, autonomous driving, and public traffic management. To enhance privacy protection, images can be encoded as features, replaced with synthetic data, or anonymized by masking human faces before being processed by the localization models.

\textbf{Data availability}. We use the publicly available CVCS, Wildtrack, and MultiviewX datasets for multi-view crowd localization and tracking. All datasets are used in accordance with their original licenses and terms of use.


\bibliographystyle{ieeetr}
\bibliography{egbib}

@String(CVPR  = {IEEE Conf. Comput. Vis. Pattern Recog.})

@String(ICCV  = {Int. Conf. Comput. Vis.})

@String(ECCV  = {Eur. Conf. Comput. Vis.})

@String(ICLR  = {Int. Conf. Learn. Represent.})

@String(BMVC  = {Brit. Mach. Vis. Conf.})

@String(AAAI  = {AAAI})

@String(IJCAI = {IJCAI})

@String(CVPR  = {CVPR})

@String(ICCV  = {ICCV})

@String(ECCV  = {ECCV})

@String(ICLR  = {ICLR})

@String(BMVC  =	{BMVC})

@inproceedings{liu2023point,
  title={Point-Query Quadtree for Crowd Counting, Localization, and More},
  author={Liu, Chengxin and Lu, Hao and Cao, Zhiguo and Liu, Tongliang},
  booktitle={Proceedings of the IEEE/CVF International Conference on Computer Vision},
  pages={1676--1685},
  year={2023}
}

@inproceedings{zhang2019wide,
  title={Wide-Area Crowd Counting via Ground-Plane Density Maps and Multi-View Fusion CNNs},
  author={Zhang, Qi and Chan, Antoni B},
  booktitle={CVPR},
  pages={8297--8306},
  year={2019}
}

@article{viola2004robust,
  title={Robust real-time face detection},
  author={Viola, Paul and Jones, Michael J},
  journal={International Journal of Computer Vision},
  volume={57},
  number={2},
  pages={137--154},
  year={2004},
  publisher={Springer}
}

@inproceedings{sabzmeydani2007detecting,
  title={Detecting pedestrians by learning shapelet features},
  author={Sabzmeydani, Payam and Mori, Greg},
  booktitle={CVPR},
  pages={1--8},
  year={2007},
  organization={IEEE}
}

@article{wu2007detection,
  title={Detection and tracking of multiple, partially occluded humans by bayesian combination of edgelet based part detectors},
  author={Wu, Bo and Nevatia, Ram},
  journal={International Journal of Computer Vision},
  volume={75},
  number={2},
  pages={247--266},
  year={2007},
  publisher={Springer}
}

@inproceedings{zhang2015cross,
  title={Cross-scene crowd counting via deep convolutional neural networks},
  author={Zhang, Cong and Li, Hongsheng and et al.},
  booktitle={CVPR},
  pages={833--841},
  year={2015}
}

@inproceedings{onoro2016towards,
  title={Towards perspective-free object counting with deep learning},
  author={Onoro-Rubio, Daniel and L{\'o}pez-Sastre, Roberto J},
  booktitle={ECCV},
  pages={615--629},
  year={2016},
  organization={Springer}
}

@inproceedings{sam2017switching,
  title={Switching convolutional neural network for crowd counting},
  author={Sam, Deepak Babu and Surya, Shiv and Babu, R Venkatesh},
  booktitle={CVPR},
  pages={4031--4039},
  year={2017}
}

@inproceedings{li2018csrnet,
  title={Csrnet: Dilated convolutional neural networks for understanding the highly congested scenes},
  author={Li, Yuhong and Zhang, Xiaofan and Chen, Deming},
  booktitle={CVPR},
  pages={1091--1100},
  year={2018}
}

@inproceedings{sindagi2017generating,
  title={Generating high-quality crowd density maps using contextual pyramid cnns},
  author={Sindagi, Vishwanath A and Patel, Vishal M},
  booktitle={ICCV},
  pages={1879--1888},
  year={2017}
}

@inproceedings{shi2019revisiting,
  title={Revisiting perspective information for efficient crowd counting},
  author={Shi, Miaojing and Yang, Zhaohui and et al.},
  booktitle={CVPR},
  pages={7279--7288},
  year={2019}
}

@inproceedings{Ge2010Crowd,
  title={Crowd Detection with a Multiview Sampler},
  author={Ge, Weina and Collins, Robert T.},
  booktitle={ECCV},
  pages={324-337},
  year={2010},
}

@inproceedings{Xu2016Multi,
  title={Multi-view People Tracking via Hierarchical Trajectory Composition},
  author={Xu, Yuanlu and Liu, Xiaobai and Liu, Yang and Zhu, Song Chun},
  booktitle={CVPR},
  pages={4256-4265},
  year={2016},
}

@article{fleuret2007multicamera,
  title={Multicamera people tracking with a probabilistic occupancy map},
  author={Fleuret, Francois and Berclaz, Jerome and Lengagne, Richard and Fua, Pascal},
  journal={IEEE Transactions on Pattern Analysis and Machine Intelligence},
  volume={30},
  number={2},
  pages={267--282},
  year={2007},
  publisher={IEEE}
}

@inproceedings{chavdarova2017deep,
  title={Deep multi-camera people detection},
  author={Chavdarova, Tatjana and Fleuret, Fran{\c{c}}ois},
  booktitle={2017 16th IEEE International Conference on Machine Learning and Applications (ICMLA)},
  pages={848--853},
  year={2017},
  organization={IEEE}
}

@inproceedings{baque2017deep,
  title={Deep occlusion reasoning for multi-camera multi-target detection},
  author={Baqu{\'e}, Pierre and Fleuret, Fran{\c{c}}ois and Fua, Pascal},
  booktitle={Proceedings of the IEEE International Conference on Computer Vision},
  pages={271--279},
  year={2017}
}

@inproceedings{hou2020multiview,
  title={Multiview detection with feature perspective transformation},
  author={Hou, Yunzhong and Zheng, Liang and Gould, Stephen},
  booktitle={Computer Vision--ECCV 2020: 16th European Conference, Glasgow, UK, August 23--28, 2020, Proceedings, Part VII 16},
  pages={1--18},
  year={2020},
  organization={Springer}
}

@inproceedings{song2021stacked,
  title={Stacked homography transformations for multi-view pedestrian detection},
  author={Song, Liangchen and Wu, Jialian and Yang, Ming and Zhang, Qian and Li, Yuan and Yuan, Junsong},
  booktitle={Proceedings of the IEEE/CVF International Conference on Computer Vision},
  pages={6049--6057},
  year={2021}
}

@inproceedings{hou2021multiview,
  title={Multiview detection with shadow transformer (and view-coherent data augmentation)},
  author={Hou, Yunzhong and Zheng, Liang},
  booktitle={Proceedings of the 29th ACM International Conference on Multimedia},
  pages={1673--1682},
  year={2021}
}

@inproceedings{qiu20223d,
  title={3D Random Occlusion and Multi-layer Projection for Deep Multi-camera Pedestrian Localization},
  author={Qiu, Rui and Xu, Ming and Yan, Yuyao and Smith, Jeremy S and Yang, Xi},
  booktitle={Computer Vision--ECCV 2022: 17th European Conference, Tel Aviv, Israel, October 23--27, 2022, Proceedings, Part X},
  pages={695--710},
  year={2022},
  organization={Springer}
}

@inproceedings{chavdarova2018wildtrack,
  title={WILDTRACK: A Multi-camera HD Dataset for Dense Unscripted Pedestrian Detection},
  author={Chavdarova, Tatjana and Baqu{\'e}, Pierre and Bouquet, St{\'e}phane and Maksai, Andrii and Jose, Cijo and Bagautdinov, Timur and Lettry, Louis and Fua, Pascal and Van Gool, Luc and Fleuret, Fran{\c{c}}ois},
  booktitle={Proceedings of the IEEE Conference on Computer Vision and Pattern Recognition},
  pages={5030--5039},
  year={2018}
}

@inproceedings{iskakov2019learnable,
  title={Learnable Triangulation of Human Pose},
  author={Iskakov, Karim and Burkov, Egor and Lempitsky, Victor and Malkov, Yury},
  booktitle={ICCV},
  year={2019}
}

@inproceedings{yan2019perspective,
  title={Perspective-guided convolution networks for crowd counting},
  author={Yan, Zhaoyi and Yuan, Yuchen and Zuo, Wangmeng and Tan, Xiao and Wang, Yezhen and Wen, Shilei and Ding, Errui},
  booktitle={Proceedings of the IEEE/CVF International Conference on Computer Vision},
  pages={952--961},
  year={2019}
}

@inproceedings{zhang2021cross,
  title={Cross-View Cross-Scene Multi-View Crowd Counting},
  author={Zhang, Qi and Lin, Wei and Chan, Antoni B},
  booktitle={Proceedings of the IEEE/CVF Conference on Computer Vision and Pattern Recognition},
  pages={557--567},
  year={2021}
}

@inproceedings{song2021rethinking,
  title={Rethinking counting and localization in crowds: A purely point-based framework},
  author={Song, Qingyu and Wang, Changan and Jiang, Zhengkai and Wang, Yabiao and Tai, Ying and Wang, Chengjie and Li, Jilin and Huang, Feiyue and Wu, Yang},
  booktitle={Proceedings of the IEEE/CVF international conference on computer vision},
  pages={3365--3374},
  year={2021}
}

@article{you2020real,
  title={Real-time 3d deep multi-camera tracking},
  author={You, Quanzeng and Jiang, Hao},
  journal={arXiv preprint arXiv:2003.11753},
  year={2020}
}

@inproceedings{zhu2020deformable,
  title={Deformable detr: Deformable transformers for end-to-end object detection},
  author={Zhu, Xizhou and Su, Weijie and Lu, Lewei and Li, Bin and Wang, Xiaogang and Dai, Jifeng},
  booktitle={Proceedings of the International Conference on Learning Representations (ICLR)},
  year={2021}
}

@inproceedings{abousamra2021localization,
  title={Localization in the crowd with topological constraints},
  author={Abousamra, Shahira and Hoai, Minh and Samaras, Dimitris and Chen, Chao},
  booktitle={Proceedings of the AAAI Conference on Artificial Intelligence},
  volume={35},
  number={2},
  pages={872--881},
  year={2021}
}

@inproceedings{wan2021generalized,
  title={A generalized loss function for crowd counting and localization},
  author={Wan, Jia and Liu, Ziquan and Chan, Antoni B},
  booktitle={Proceedings of the IEEE/CVF Conference on Computer Vision and Pattern Recognition},
  pages={1974--1983},
  year={2021}
}

@inproceedings{ma2019bayesian,
  title={Bayesian loss for crowd count estimation with point supervision},
  author={Ma, Zhiheng and Wei, Xing and Hong, Xiaopeng and Gong, Yihong},
  booktitle={Proceedings of the IEEE/CVF international conference on computer vision},
  pages={6142--6151},
  year={2019}
}

@article{wang2020distribution,
  title={Distribution matching for crowd counting},
  author={Wang, Boyu and Liu, Huidong and Samaras, Dimitris and Nguyen, Minh Hoai},
  journal={Advances in neural information processing systems},
  volume={33},
  pages={1595--1607},
  year={2020}
}

@inproceedings{ma2021learning,
  title={Learning to count via unbalanced optimal transport},
  author={Ma, Zhiheng and Wei, Xing and Hong, Xiaopeng and Lin, Hui and Qiu, Yunfeng and Gong, Yihong},
  booktitle={Proceedings of the AAAI Conference on Artificial Intelligence},
  volume={35},
  number={3},
  pages={2319--2327},
  year={2021}
}

@inproceedings{liu2021bipartite,
  title={Bipartite Matching for Crowd Counting with Point Supervision.},
  author={Liu, Hao and Zhao, Qiang and Ma, Yike and Dai, Feng},
  booktitle={IJCAI},
  pages={860--866},
  year={2021}
}

@inproceedings{laradji2018blobs,
  title={Where are the blobs: Counting by localization with point supervision},
  author={Laradji, Issam H and Rostamzadeh, Negar and Pinheiro, Pedro O and Vazquez, David and Schmidt, Mark},
  booktitle={Proceedings of the european conference on computer vision (ECCV)},
  pages={547--562},
  year={2018}
}

@InProceedings{Lin_2023_CVPR,
    author    = {Lin, Wei and Chan, Antoni B.},
    title     = {Optimal Transport Minimization: Crowd Localization on Density Maps for Semi-Supervised Counting},
    booktitle = {Proceedings of the IEEE/CVF Conference on Computer Vision and Pattern Recognition (CVPR)},
    month     = {June},
    year      = {2023},
    pages     = {21663-21673}
}

@article{liang2022focal,
  title={Focal inverse distance transform maps for crowd localization},
  author={Liang, Dingkang and Xu, Wei and Zhu, Yingying and Zhou, Yu},
  journal={IEEE Transactions on Multimedia},
  year={2022},
  publisher={IEEE}
}

@inproceedings{yan2022multiview,
  title={Multiview transformers for video recognition},
  author={Yan, Shen and Xiong, Xuehan and Arnab, Anurag and Lu, Zhichao and Zhang, Mi and Sun, Chen and Schmid, Cordelia},
  booktitle={Proceedings of the IEEE/CVF conference on computer vision and pattern recognition},
  pages={3333--3343},
  year={2022}
}

@article{peyre2017computational,
  title={Computational optimal transport},
  author={Peyr{\'e}, Gabriel and Cuturi, Marco and others},
  journal={Center for Research in Economics and Statistics Working Papers},
  number={2017-86},
  year={2017}
}

@InProceedings{Cheng_2023_ICCV,
    author    = {Cheng, Cheng-Che and Qiu, Min-Xuan and Chiang, Chen-Kuo and Lai, Shang-Hong},
    title     = {ReST: A Reconfigurable Spatial-Temporal Graph Model for Multi-Camera Multi-Object Tracking},
    booktitle = {Proceedings of the IEEE/CVF International Conference on Computer Vision (ICCV)},
    month     = {October},
    year      = {2023},
    pages     = {10051-10060}
}

@InProceedings{Teepe_2024_WACV,
    author    = {Teepe, Torben and Wolters, Philipp and Gilg, Johannes and Herzog, Fabian and Rigoll, Gerhard},
    title     = {EarlyBird: Early-Fusion for Multi-View Tracking in the Bird's Eye View},
    booktitle = {Proceedings of the IEEE/CVF Winter Conference on Applications of Computer Vision (WACV) Workshops},
    month     = {January},
    year      = {2024},
    pages     = {102-111}
}

@InProceedings{Jeon_2023_CVPR,
    author    = {Jeon, Yuntae and Tran, Dai Quoc and Park, Minsoo and Park, Seunghee},
    title     = {Leveraging Future Trajectory Prediction for Multi-Camera People Tracking},
    booktitle = {Proceedings of the IEEE/CVF Conference on Computer Vision and Pattern Recognition (CVPR) Workshops},
    month     = {June},
    year      = {2023},
    pages     = {5399-5408}
}

@InProceedings{Kim_2024_CVPR,
    author    = {Kim, Jeongho and Shin, Wooksu and Park, Hancheol and Choi, Donghyuk},
    title     = {Cluster Self-Refinement for Enhanced Online Multi-Camera People Tracking},
    booktitle = {Proceedings of the IEEE/CVF Conference on Computer Vision and Pattern Recognition (CVPR) Workshops},
    month     = {June},
    year      = {2024},
    pages     = {7190-7197}
}

@inproceedings{niculescu2025mctr,
  title={MCTR: Multi camera tracking transformer},
  author={Niculescu-Mizil, Alexandru and Patel, Deep and Melvin, Iain},
  booktitle={Proceedings of the Winter Conference on Applications of Computer Vision},
  pages={874--884},
  year={2025}
}

@inproceedings{cherdchusakulchai2024online,
  title={Online Multi-camera People Tracking with Spatial-temporal Mechanism and Anchor-feature Hierarchical Clustering},
  author={Cherdchusakulchai, Riu and Phimsiri, Sasin and Trairattanapa, Visarut and Tungjitnob, Suchat and Kudisthalert, Wasu and Kiawjak, Pornprom and Thamwiwatthana, Ek and Borisuitsawat, Phawat and Tosawadi, Teepakorn and Choppradi, Pakcheera and others},
  booktitle={Proceedings of the IEEE/CVF Conference on Computer Vision and Pattern Recognition},
  pages={7198--7207},
  year={2024}
}

@inproceedings{carion2020end,
  title={End-to-end object detection with transformers},
  author={Carion, Nicolas and Massa, Francisco and Synnaeve, Gabriel and Usunier, Nicolas and Kirillov, Alexander and Zagoruyko, Sergey},
  booktitle={European conference on computer vision},
  pages={213--229},
  year={2020},
  organization={Springer}
}

@InProceedings{Kim_2023_CVPR,
    author    = {Jeongho Kim, Wooksu Shin, Hancheol Park and Jongwon Baek},
    title     = {Addressing the Occlusion Problem in Multi-Camera People Tracking with Human Pose Estimation},
    booktitle = {Proceedings of the IEEE/CVF Conference on Computer Vision and Pattern Recognition (CVPR) Workshops},
    month     = {June},
    year      = {2023},
}

@article{kalman1960new,
  title={A new approach to linear filtering and prediction problems},
  author={Kalman, Rudolph Emil},
  year={1960}
}

@article{ong2020bayesian,
  title={A Bayesian filter for multi-view 3D multi-object tracking with occlusion handling},
  author={Ong, Jonah and Vo, Ba-Tuong and Vo, Ba-Ngu and Kim, Du Yong and Nordholm, Sven},
  journal={IEEE Transactions on Pattern Analysis and Machine Intelligence},
  volume={44},
  number={5},
  pages={2246--2263},
  year={2020},
  publisher={IEEE}
}

@inproceedings{engilberge2023multi,
  title={Multi-view tracking using weakly supervised human motion prediction},
  author={Engilberge, Martin and Liu, Weizhe and Fua, Pascal},
  booktitle={Proceedings of the IEEE/CVF Winter Conference on Applications of Computer Vision},
  pages={1582--1592},
  year={2023}
}

@inproceedings{he2023lef,
  title={Lef: Late-to-early temporal fusion for lidar 3d object detection},
  author={He, Tong and Sun, Pei and Leng, Zhaoqi and Liu, Chenxi and Anguelov, Dragomir and Tan, Mingxing},
  booktitle={2023 IEEE/RSJ International Conference on Intelligent Robots and Systems (IROS)},
  pages={1637--1644},
  year={2023},
  organization={IEEE}
}

@inproceedings{MVOT24,
title={Mahalanobis Distance-based Multi-view Optimal Transport for Multi-view Crowd Localization},
author={Qi Zhang and Kaiyi Zhang and Antoni B. Chan and Hui Huang},
booktitle={ECCV},

volume={15062},
pages={19--36},
year={2024},
}

@inproceedings{yamane2025msmvd,
  title={MSMVD: Exploiting Multi-scale Image Features via Multi-scale BEV Features for Multi-view Pedestrian Detection},
  author={Yamane, Taiga and Suzuki, Satoshi and Masumura, Ryo and Orihashi, Shota and Tanaka, Tomohiro and Ihori, Mana and Makishima, Naoki and Kawata, Naotaka},
  booktitle={Proceedings of the British Machine Vision Conference (BMVC)},
  year={2025}
}

@inproceedings{daryani2025camuvid,
  title={CaMuViD: Calibration-Free Multi-View Detection},
  author={Daryani, Amir Etefaghi and Bhutta, M and Hernandez, Byron and Medeiros, Henry},
  booktitle={Proceedings of the Computer Vision and Pattern Recognition Conference},
  pages={1220--1229},
  year={2025}
}

@inproceedings{fan2025all,
  title={All-Day Multi-Camera Multi-Target Tracking},
  author={Fan, Huijie and Qiao, Yu and Zhen, Yihao and Zhao, Tinghui and Fan, Baojie and Wang, Qiang},
  booktitle={Proceedings of the Computer Vision and Pattern Recognition Conference},
  pages={16892--16901},
  year={2025}
}

@InProceedings{Teepe_2024_CVPR,
    author    = {Teepe, Torben and Wolters, Philipp and Gilg, Johannes and Herzog, Fabian and Rigoll, Gerhard},
    title     = {Lifting Multi-View Detection and Tracking to the Bird's Eye View},
    booktitle = {Proceedings of the IEEE/CVF Conference on Computer Vision and Pattern Recognition (CVPR) Workshops},
    month     = {June},
    year      = {2024},
    pages     = {667-676}
}

@INPROCEEDINGS{11018739,
  author={Dakic, Kosta and Thilakarathna, Kanchana and Calheiros, Rodrigo N. and Lim, Teng Joon},
  booktitle={2025 IEEE International Conference on Pervasive Computing and Communications (PerCom)}, 
  title={Resource-Efficient Multiview Perception: Integrating Semantic Masking with Masked Autoencoders}, 
  year={2025},
  volume={},
  number={},
  pages={145-151},
  doi={10.1109/PerCom64205.2025.00034}}

@article{yang2024end,
  title={An end-to-end tracking framework via multi-view and temporal feature aggregation},
  author={Yang, Yihan and Xu, Ming and Ralph, Jason F and Ling, Yuchen and Pan, Xiaonan},
  journal={Computer Vision and Image Understanding},
  volume={249},
  pages={104203},
  year={2024},
  publisher={Elsevier}
}

@article{khosla2020supervised,
  title={Supervised contrastive learning},
  author={Khosla, Prannay and Teterwak, Piotr and Wang, Chen and Sarna, Aaron and Tian, Yonglong and Isola, Phillip and Maschinot, Aaron and Liu, Ce and Krishnan, Dilip},
  journal={Advances in neural information processing systems},
  volume={33},
  pages={18661--18673},
  year={2020}
}

@ARTICLE{10742444,
  author={Zhang, Wei and Li, Jiaming and Xia, Meng and Gao, Xu and Tan, Xiao and Shi, Yifeng and Huang, Zhenhua and Li, Guanbin},
  journal={IEEE Transactions on Pattern Analysis and Machine Intelligence}, 
  title={OffsetNet: Towards Efficient Multiple Object Tracking, Detection, and Segmentation}, 
  year={2025},
  volume={47},
  number={2},
  pages={949-960},
  doi={10.1109/TPAMI.2024.3485644}}

@article{ZHANG2026113396,
title = {A novel geometry-aware spatio-temporal network for multi-view video feature learning},
journal = {Pattern Recognition},
volume = {178},
pages = {113396},
year = {2026},
issn = {0031-3203},
doi = {https://doi.org/10.1016/j.patcog.2026.113396},
url = {https://www.sciencedirect.com/science/article/pii/S0031320326003614},
author = {Yue Zhang and Mai Xu and Lai Jiang and Xin Deng and Si Liu}
}

@article{ji2026view,
  title={View-centric multi-object tracking with homographic matching in moving uav},
  author={Ji, Deyi and Zhu, Lanyun and Gao, Siqi and Zhu, Qi and Zhao, Yiru and Xu, Peng and Ding, Yue and Lu, Hongtao and Ye, Jieping and Wu, Feng and others},
  journal={IEEE Transactions on Geoscience and Remote Sensing},
  year={2026},
  publisher={IEEE}
}

@article{anderson2026model,
  title={Model Optimization for Multi-Camera 3D Detection and Tracking},
  author={Anderson, Ethan and Silva, Justin and Zheng, Kyle and Pusegaonkar, Sameer and Wang, Yizhou and Tang, Zheng and Biswas, Sujit},
  journal={arXiv preprint arXiv:2602.00450},
  year={2026}
}

@article{kuhn1955hungarian,
  title={The Hungarian method for the assignment problem},
  author={Kuhn, Harold W},
  journal={Naval research logistics quarterly},
  volume={2},
  number={1-2},
  pages={83--97},
  year={1955},
  publisher={Wiley Online Library}
}

@article{bernardin2008evaluating,
  title={Evaluating multiple object tracking performance: the clear mot metrics},
  author={Bernardin, Keni and Stiefelhagen, Rainer},
  journal={EURASIP Journal on Image and Video Processing},
  volume={2008},
  number={1},
  pages={246309},
  year={2008},
  publisher={Springer}
}

@InProceedings{Yamane_2025_ICCV,
    author    = {Yamane, Taiga and Masumura, Ryo and Suzuki, Satoshi and Orihashi, Shota},
    title     = {MVTrajecter: Multi-View Pedestrian Tracking with Trajectory Motion Cost and Trajectory Appearance Cost},
    booktitle = {Proceedings of the IEEE/CVF International Conference on Computer Vision (ICCV)},
    month     = {October},
    year      = {2025},
    pages     = {13270-13280}
}

\end{document}